\documentclass[10pt,twocolumn,letterpaper]{article}

\usepackage[T1]{fontenc}    
\usepackage{booktabs}       
\usepackage{amsfonts}       
\usepackage{nicefrac}       
\usepackage{microtype}      
\usepackage[table]{xcolor}         
\usepackage{makecell}
\usepackage{listings}
\usepackage{float}
\usepackage{pifont}

\usepackage{graphicx}
\usepackage{booktabs}
\usepackage{indentfirst}
\usepackage{xcolor}
\newcommand{\xmark}{\ding{55}}%

\usepackage[accsupp]{axessibility}  

\definecolor{codegreen}{rgb}{0,0.6,0}
\definecolor{codegray}{rgb}{0.5,0.5,0.5}
\definecolor{codepurple}{rgb}{0.58,0,0.82}
\definecolor{backcolour}{rgb}{0.95,0.95,0.92}
\definecolor{purple}{rgb}{0.56,0.27,0.68}
\definecolor{red}{rgb}{0.95,0.4,0.4}
\definecolor{purered}{rgb}{1,0,0}
\definecolor{blue}{rgb}{0.4,0.4,0.95}
\definecolor{darkblue}{rgb}{0,0,0.8}
\definecolor{grey}{rgb}{0.6,0.6,0.6}
\definecolor{col1}{RGB}{232, 161, 148}
\definecolor{col11}{RGB}{255, 228, 228}
\definecolor{col2}{RGB}{148, 187, 232}
\definecolor{col33}{RGB}{206, 239, 255}
\definecolor{col3}{RGB}{233, 255, 245}
\definecolor{lightgrey}{rgb}{0.85,0.85,0.85}
\definecolor{lightlightgrey}{rgb}{0.9,0.9,0.9}
\definecolor{verylightBG}{rgb}{0.9,0.99,0.99}
\definecolor{darkgreen}{rgb}{0., 0.85, 0.5}

\lstdefinestyle{pythonstyle}{
    backgroundcolor=\color{backcolour},
    commentstyle=\color{codegreen},
    keywordstyle=\color{magenta},
    numberstyle=\tiny\color{codegray},
    stringstyle=\color{codepurple},
    basicstyle=\ttfamily\footnotesize,
    breakatwhitespace=false,
    breaklines=true,
    captionpos=b,
    keepspaces=true,
    numbers=left,
    numbersep=5pt,
    showspaces=false,
    showstringspaces=false,
    showtabs=false,
    tabsize=2,
    frame=single
}

\newcommand\greybox[1]{%
  \vskip\baselineskip%
  \par\noindent\colorbox{lightgray}{%
    \begin{minipage}{\linewidth}#1\end{minipage}%
  }%
  \vskip\baselineskip%
}

 \usepackage[pagenumbers]{cvpr} 

\definecolor{cvprblue}{rgb}{0.21,0.49,0.74}
\usepackage[pagebackref,breaklinks,colorlinks,allcolors=cvprblue]{hyperref}

\def\paperID{xxxx} 
\def\confName{CVPR}
\def\confYear{2026}

\title{RefAV: Towards Planning-Centric Scenario Mining}

\author{Cainan Davidson, Deva Ramanan, Neehar Peri \\
Carnegie Mellon University
}

\begin{document}
\maketitle

\begin{abstract}
Autonomous Vehicles (AVs) collect and pseudo-label terabytes of multi-modal data localized to HD maps during normal fleet testing. However, identifying interesting and safety-critical scenarios from uncurated driving logs remains a significant challenge. Traditional scenario mining techniques are error-prone and prohibitively time-consuming, often relying on hand-crafted structured queries. In this work, we revisit spatio-temporal scenario mining through the lens of recent vision-language models (VLMs) to detect whether a described scenario occurs in a driving log and, if so, precisely localize it in both time and space. To address this problem, we introduce RefAV, a large-scale dataset of $10,000$ diverse natural language queries that describe complex multi-agent interactions relevant to motion planning derived from $1000$ driving logs in the Argoverse 2 Sensor dataset. We evaluate several referential multi-object trackers and present an empirical analysis of our baselines. Notably, we find that naively repurposing off-the-shelf VLMs yields poor performance, suggesting that scenario mining presents unique challenges. Lastly, we discuss our recently held competition and share insights from the community. Our code and dataset are available on \href{https://github.com/CainanD/RefAV}{Github} and \href{https://argoverse.github.io/user-guide/tasks/scenario_mining.html}{Argoverse}. 
\end{abstract}

\begin{figure*}[t]
    \centering
    \includegraphics[width=\linewidth,trim={0 0cm 0 0cm}, clip]{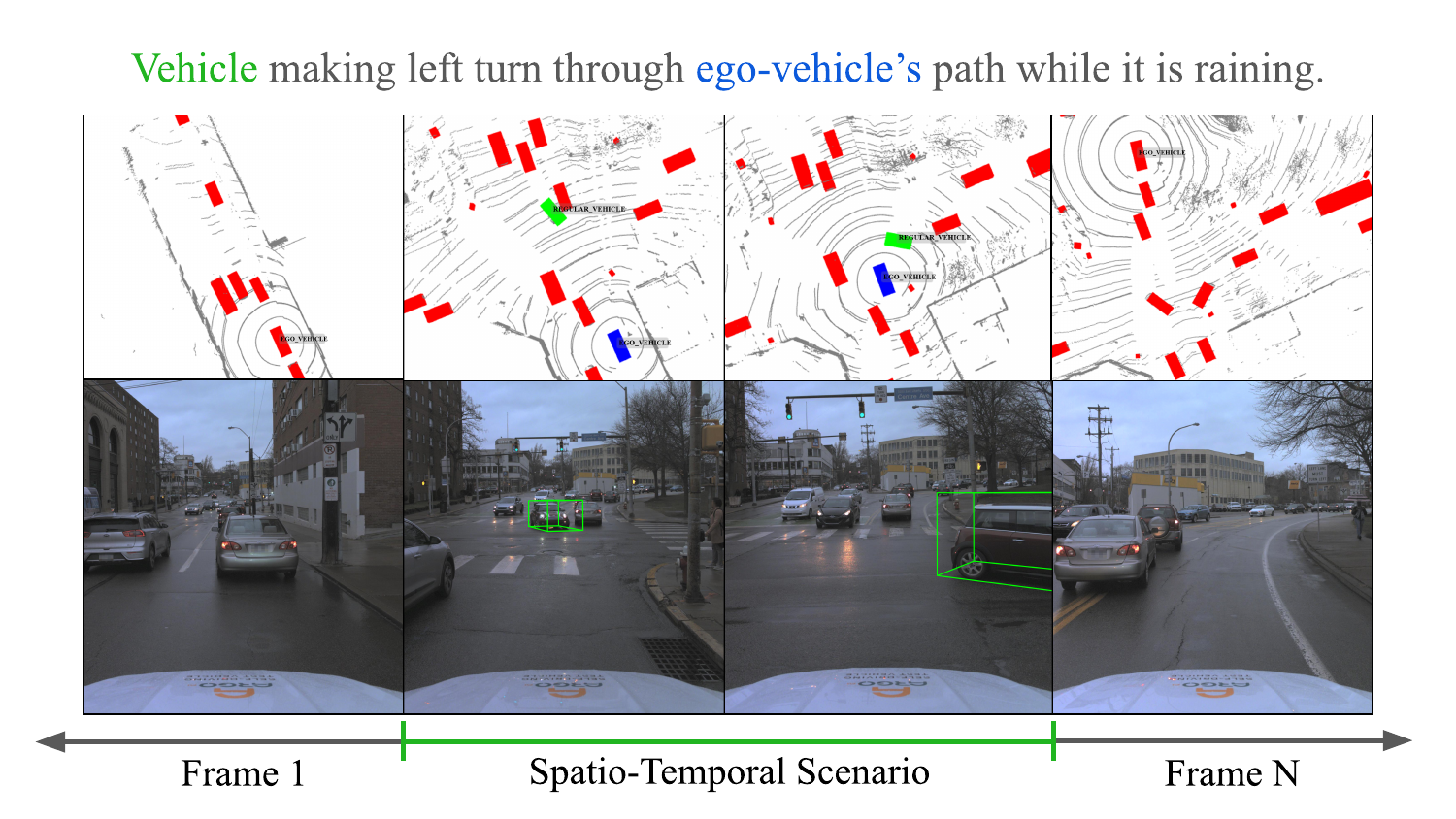}
    \caption{{\bf Scenario Mining Problem Setup.} Given a natural language prompt such as {\tt vehicle making left turn through ego-vehicle's path while it is raining}, our problem setup requires models to determine whether the described scenario occurs within a 20-second driving log, and if so, precisely localize the referred object in 3D space and time from raw sensor data (LiDAR, 360$^\circ$ ring cameras, and HD maps). Based on the example above, a VLM should localize the start and end timestamps and 3D location of the red Mini Cooper executing a ``Pittsburgh left'' through the ego-vehicle's path with a 3D track. Notably, the ``Pittsburgh left'' is a regional driving practice where a driver quickly makes a left turn before oncoming traffic proceeds. Although common in Pittsburgh, this maneuver is technically illegal. Therefore, we argue that scenario mining is critical for validating end-to-end autonomy in order to build a comprehensive safety case. Note that \textcolor{green!80!black}{\tt referred objects} are shown in green, \textcolor{blue}{\tt related objects} in blue, and \textcolor{red}{\tt other objects} in red. 
    }
    \label{fig:teaser}
\end{figure*}

\section{Introduction}
\label{sec:intro}


Autonomous Vehicle (AV) deployment on public roads has increased significantly in recent years, with Waymo completing more than a million rides per week as of February 2025 \cite{Korosec2025}. Despite the maturity of such autonomous ride-hailing services, AVs can still sometimes require manual interventions \cite{korosec2025waymo, mihalascu2023cruise} and can be prone to accidents~\cite{thadani2023cruise}. 
Although data-driven simulators are an integral component in establishing a safety case \cite{karnchanachari2024towards, gulino2023waymax, vasudevan2024planning, yang2023unisim}, validating end-to-end autonomy with real-world operational data remains a critical part of the testing stack. However, identifying interesting and safety-critical scenarios from uncurated real-world driving logs is akin to finding a "needle in a haystack" due to the scale of data collected during normal fleet operations. In this paper, we revisit the task of spatio-temporal scenario mining with recent vision-language models (VLMs) to identify interesting multi-agent interactions using natural language (cf. Fig. \ref{fig:teaser}). 

{\bf Status Quo.} Although language-based 3D scene understanding has been extensively studied in the context of referential multi-object tracking (RMOT) \cite{wu2023referkitti, wu2023nuprompt, li2025nugrounding}, multi-modal visual question answering (VQA) \cite{qian2024nuscenesqa, Inoue_2024_WACV, tian2025nuscenes}, and VLM-based motion planning \cite{sima2024drivelm, mao2023gpt, shao2024lmdrive, wang2024omnidrive}, we argue that spatiotemporal scenario mining presents unique challenges. RMOT extends referential grounding by associating referred objects over time. However, unlike RMOT, scenario mining does not guarantee that referred objects exist in a given driving log. Next, multi-modal VQA extends VQA with additional visual modalities like LiDAR. Although spatio-temporal scenario mining also leverages multi-modal input (e.g. RGB, LiDAR and HD maps), we require that methods output 3D tracks instead of text-based answers. Lastly, VLM-based motion planners directly estimate future ego-vehicle waypoints based on high-level language instructions and past sensor measurements. In contrast, scenario mining methods can reason over the full driving log to identify interactions between non-ego vehicles. Concretely, spatio-temporal scenario mining requires identifying if a described scenario occurs in a driving log from raw sensor measurements (e.g. RGB and LiDAR), and if so, precisely localizing {\em all} referred objects in both time and space with 3D tracks. To support this task, we propose RefAV, a large-scale dataset of $10,000$ diverse natural language queries designed to evaluate a model's ability to find a visual ``needle in a haystack''. 

{\bf RefAV Scenario Mining Dataset.} We repurpose the Argoverse 2 (AV2) Sensor dataset \cite{wilson2023argoverse}, which contains $1,000$ driving logs with synchronized LiDAR, 360$^\circ$ ring cameras, HD maps, and 3D track annotations for $30$ categories. We curate a set of $10,000$ natural language prompts that describe interesting (and often rare) scenarios (cf. Fig. \ref{fig:dataset_creation} and Fig. \ref{fig:dataset_examples}) using a combination of manual annotations and procedural generation using large language models (LLMs). Notably, AV2 annotates ground truth tracks at 10 Hz, which allows for fine-grained motion understanding. In contrast, prior referential multi-object tracking benchmarks (the established task most similar to spatio-temporal scenario mining) use nuScenes \cite{caesar2020nuscenes}, which annotates ground truth tracks at 2 Hz (cf. Table \ref{tab:other_datasets}). Argoverse's higher temporal resolution uniquely allows RefAV to evaluate dynamic multi-agent interactions (unlike prior work that primarily evaluates referential expressions based on static attributes like vehicle color or relative heading). Interestingly, we find that simply repurposing VLMs for scenario mining yields poor performance.

{\bf Referential Tracking by Program Synthesis.} Although prior referential trackers and VLMs achieve reasonable performance with simple referential prompts (e.g. {\tt find all cars}), we find that such methods struggle with compositional reasoning and motion understanding (e.g. {\tt find all cars accelerating while changing lanes}). To address this problem, we propose Referential Tracking by Program Synthesis (RefProg), a modular approach that combines off-the-shelf 3D tracks with LLMs. Inspired by recent work in program synthesis \cite{surismenon2023vipergpt,gupta2023visual, marsili2025visual}, our method uses an LLM to break down complex referential expressions into simpler compositional actions. Specifically, we define an API to describe hand-crafted atomic functions (e.g., {\tt turning}, {\tt accelerating}, {\tt changing lanes}) and use an LLM to synthesize a program that composes these atomic actions corresponding to the natural language prompt. We then execute the generated program to filter off-the-shelf tracks to identify the subset of tracks that best match the described scenario (cf. Fig. \ref{fig:baseline}).

{\bf Contributions.} We present three major contributions. First, we introduce RefAV, a large-scale dataset designed to evaluate VLMs on 3D scene understanding and spatio-temporal localization.  Our extensive experiments highlight the limitations of current methods, and demonstrate the effectiveness of our proposed program synthesis-based approach. Lastly, we highlight the results of our recent  \href{https://eval.ai/web/challenges/challenge-page/2469/evaluation}{CVPR 2025 challenge} hosted in conjunction with the \href{https://cvpr2025.wad.vision/}{Workshop on Autonomous Driving}. 

\section{Related Works}
\label{sec:related-works}

\textbf{Referential Grounding and Tracking} are long-standing challenges in vision-language understanding. Early datasets like RefCOCO, RefCOCO+, and RefCOCOg \cite{refcoco, refcoco+, refcocog} helped popularize the task, which was later adapted for autonomous driving \cite{deruyttere2019talk2car, Vasudevan_2018_CVPR}. Although prior work focused on single-frame visual grounding, ReferKITTI \cite{wu2023referkitti} formalizes the problem of 2D referential multi-object tracking (RMOT), which extends referential object detection by associating referred objects over time. ReferKITTI-v2 \cite{zhang2024bootstrapping} expands ReferKITTI's manual annotations using large language models (LLMs) to improve prompt diversity. Further, LaMOT \cite{li2024lamot} consolidates 1,660 sequences from four tracking datasets to create a large-scale unified RMOT benchmark. 

Referential tracking methods can be broadly classified into two-stage (e.g. separately track and filter predictions based on natural language cues) and end-to-end approaches. iKUN \cite{du2024ikun} proposes a plug-and-play knowledge unification module to facilitate language grounding for off-the-shelf trackers. Similarly, ReferGPT \cite{chamiti2025refergpt} presents a zero-shot method for filtering off-the-shelf 3D tracks using CLIP \cite{clip} scores and fuzzy matching.  In contrast, JointNLT \cite{zhou2023joint} proposes a unified visual grounding and tracking framework, and MMTrack \cite{mmtrack} reformulates tracking as token generation. OVLM \cite{patzold2025leveraging} introduces a memory-aware model for RMOT, while ReferFormer \cite{wu2022language} leverages language-conditioned object queries. More recently, nuPrompt \cite{wu2023nuprompt} extends 2D referential tracking to 3D with RGB input. Different from nuPrompt, RefAV addresses 3D {\em multi-modal} referential tracking and dynamic scene understanding.

\textbf{Visual Question Answering with VLMs} has improved significantly in recent years due to large-scale multi-modal pre-training. Vision-language models (VLMs) like LLaVA \cite{liu2023llava, liu2023improvedllava}, BLIP-2 \cite{blip2}, and Qwen2.5-VL \cite{bai2025qwen2.5vl} show strong generalization across diverse domains \cite{balanced_vqa_v2, tian2025nuscenes, qian2024nuscenesqa}. However, state-of-the-art models still struggle with spatial reasoning and grounding \cite{madan2024revisiting,robicheaux2025roboflow100}. To address this issue, SpatialRGPT \cite{cheng2024spatialrgpt} and SpatialVLM \cite{chen2024spatialvlm} generate large-scale monocular depth pseudo-labels to improve 3D reasoning. Despite the effectiveness of such methods, zero-shot prompting yields poor compositional reasoning performance. Instead, VisProg \cite{gupta2023visual} and ViperGPT \cite{surismenon2023vipergpt} use LLMs to generate executable code and use tools like OwlViT \cite{minderer2022simple}, CLIP \cite{clip}, MiDaS \cite{ranftl2020towards}, and GLIP \cite{li2022grounded} for spatial understanding. However, existing program synthesis approaches primarily focus on single-frame reasoning. In contrast, we are interested in understanding dynamic multi-agent interactions from video sequences. We take inspiration from existing methods and apply similar program synthesis-based approaches for spatio-temporal scenario mining. 

\textbf{Vision-Language Models for 3D Understanding} have been extensively explored in the context of representation learning, open-vocabulary 3D perception, and end-to-end driving.  SLidR \cite{sautier2022image} distills 2D features into 3D point clouds for cross-modal learning. SEAL \cite{liu2023segment} incorporates SAM \cite{kirillov2023segment} to produce class-agnostic 3D segments, while SA3D \cite{cen2024segment} leverages SAM and NeRFs for object segmentation. Similarly, recent work like Anything-3D \cite{shen2023anything} and 3D-Box-Segment-Anything integrate VLMs and 3D detectors (e.g., VoxelNeXt \cite{chen2023voxelnext}) for interactive reconstruction and labeling. VLMs have been used extensively in open-vocabulary perception for autonomous driving. UP-VL \cite{najibi2023unsupervised} distills CLIP features into LiDAR data to generate amodal cuboids. Recent work uses 2D VLMs to generate 3D pseudo-labels for open-vocabulary perception, enabling zero-shot LiDAR panoptic segmentation \cite{sal2024eccv, takmaz2025cal} and 3D object detection \cite{khurana2024shelf, peri2022towards, ma2023long}. While traditional grounding methods \cite{wu2023nuprompt} struggle with complex instructions, multi-modal LLMs \cite{mao2023gpt, shao2024lmdrive, cho2024language} demonstrate impressive visual understanding. However, such methods \cite{bai20243d, wang2024omnidrive, ding2024holistic} typically produce scene-level analysis for end-to-end driving rather than precise instance localization. In contrast, our framework combines language grounding with precise geometric localization from offline 3D perception methods for more accurate vision-language reasoning.

\section{RefAV: Scenario Mining with Natural Language Descriptions}
\label{sec:method}

In this section, we present our approach for curating $10,000$ natural language prompts (Sec \ref{ssec:dataset_creation}) and describe five zero-shot scenario mining baselines (Sec \ref{ssec:baselines}).

\begin{figure*}[t]
    \centering
    \includegraphics[width=\linewidth,trim={0 7cm 0 7cm}, clip]{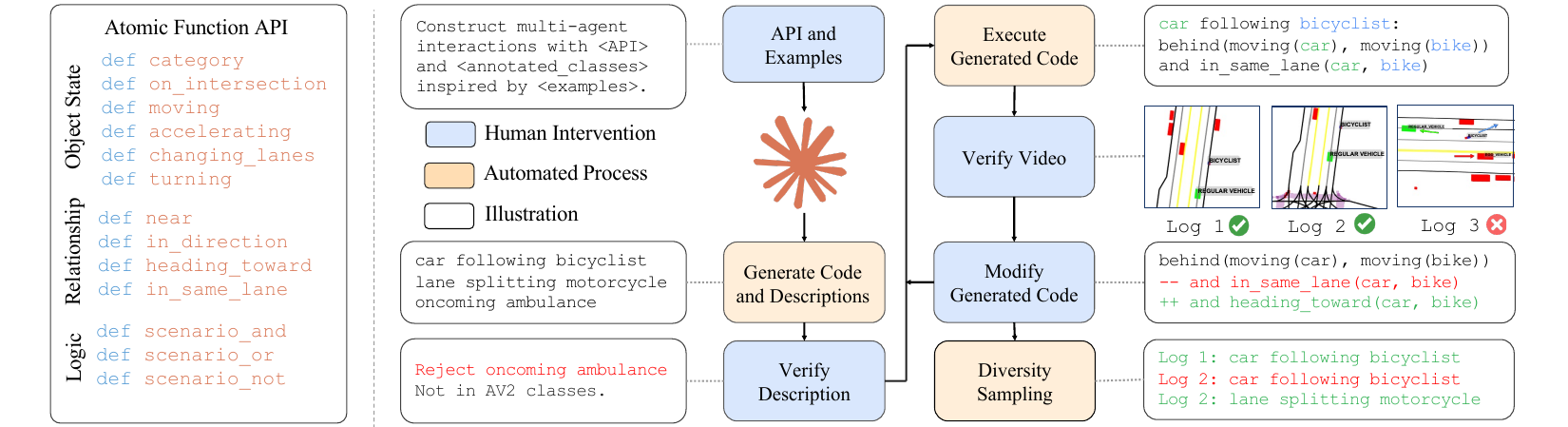}
    \caption{{\bf RefAV Dataset Creation.} First, we define a set of 28 atomic functions that identify the state of an object track, its relationship with other objects (stored in an underlying scene graph), and a set of boolean logical operators to support function composition. Next, we prompt an LLM to permute these atomic functions and generate a program and corresponding natural language description.  Finally, we execute the generated code on ground-truth tracks and visualize the referred object track to manually verify that the program output matches the natural language prompt. Code that generates an incorrect video is modified by an annotator and re-executed. We sample valid programs to maximize scenario diversity in our dataset. }
    \label{fig:dataset_creation}
\end{figure*}

\subsection{Creating RefAV}
\label{ssec:dataset_creation}

Unlike prior benchmarks that primarily evaluate referential expressions based on static attributes like vehicle color or relative heading (e.g. {\tt find the red car to the left}), we focus on mining interesting and safety-critical scenarios relevant for motion planning (cf. Fig. \ref{fig:teaser}). We take inspiration from recent planning benchmarks like nuPlan \cite{karnchanachari2024towards} to identify such planning-centric scenarios. In particular, nuPlan introduces a set of \href{https://github.com/motional/nuplan-devkit/blob/master/nuplan/planning/script/config/common/scenario_builder/scenario_mapping/nuplan_scenario_mapping.yaml}{80 scenarios} considered relevant for safe motion planning. We use this as a template to identify similar scenarios within AV2.

 \textbf{Why Use the Argoverse 2 (AV2) Sensor Dataset?} 
 The AV2 sensor dataset contains $1000$ 15 -- 20 second logs with synchronized sensor measurements from seven ring cameras and two LiDAR sensors. Moreover, the dataset includes HD maps with lane markings, crosswalk polygons, and various lane types (e.g. vehicle, bus, and bike lanes). Notably, AV2 annotates 30 object categories with track-level annotations at 10 Hz. We choose to build RefAV on top of AV2 because prior datasets like nuScenes \cite{caesar2020nuscenes} only annotate ground-truth tracks at a lower temporal resolution (e.g. 2Hz vs. 10Hz), making it more difficult to extract fine-grained motion. We ablate this in the Appendix \ref{sec:sampling_rate}. In addition, KITTI \cite{geiger2012we} and Waymo Open Dataset \cite{sun2020scalability} only label a limited number of categories (e.g. {\tt car}, {\tt pedestrian}, {\tt bicycle}) and do not have HD maps, making it difficult to evaluate diverse multi-agent interactions. Although AV2 annotates objects up to 150m away from the ego vehicle, we clip all object tracks at 50m. We find that current 3D perception models struggle with long-range detection and tracking \cite{peri2023empirical}, suggesting that the community is not ready to address spatio-temporal scenario mining at range. 

\textbf{LLM-based Procedural Scenario Generation.} 
Our key insight is that many complex scenarios (e.g. {\tt find all cars accelerating while changing lanes}) can be broken down into simpler atomic actions (e.g. {\tt find cars}, {\tt accelerating}, and {\tt changing lanes}), and a large set of such atomic actions can be composed to generate new diverse scenarios. To this end, we first define 28 atomic functions based on nuPlan's list of planning scenarios (cf. Figure \ref{fig:dataset_creation} left). We include a full API listing in the Appendix \ref{sec:api_listing}. To generate a new scenario definition, we provide an LLM (in our case, Claude 3.7 Sonnet) with the full API listing, along with in-context examples of real compositions. We prompt the LLM to generate permutations of the atomic functions and describe the generated code with a natural language description (cf. Figure \ref{fig:dataset_creation} right). We execute the generated code to filter ground-truth tracks from all $1000$ logs to identify true positive matches. We aggregate all true positive log-prompt matches (>50K) and sample 8,000 true positive log-prompt pairs that maximize scenario diversity. Lastly, we randomly sample 2,000 true negative log-prompt pairs. Our automatic scenario generation pipeline has an average success rate of 70\%. 

\textbf{Verifying Procedurally Generated Scenarios.}
Although LLM-based procedural generation allows us to generate diverse scenarios at scale, this process is not perfect and requires extensive manual validation (cf. Figure \ref{fig:dataset_creation} right). First, we verify that the generated descriptions match the generated code.  We then manually review video clips of true positive log-prompt matches to ensure that the generated natural language description accurately describes the localized tracks. Notably, we find two common error modes. First, the LLM often defines the prototypical case of a scenario, but misses or incorrectly includes edge cases due to under or overspecification.  For example, given a description {\tt car following a bicyclist}, the generated code identifies instances where a moving car is behind a moving bike while traveling in the same road lane. This definition includes the false positive edge case where the car and bicyclist are traveling in opposite directions.  Second, we find that LLMs often reverse the relationship of \texttt{referred} and {\tt related} objects. For example, given the description {\tt bicycles in front of the car}, the LLM generates code that corresponds to {\tt car behind the bicycles}. Based on the videos, human annotators make the necessary changes to the scenario definition, and the modified code is re-executed across all logs. It takes about 3 minutes to verify and edit each scenario definition.  In total, we spent 200 hours verifying scenario definitions.
\begin{figure*}[t]
    \centering
    \includegraphics[width=0.9\linewidth, clip, trim=0cm 3.1cm 0cm 3.5cm]{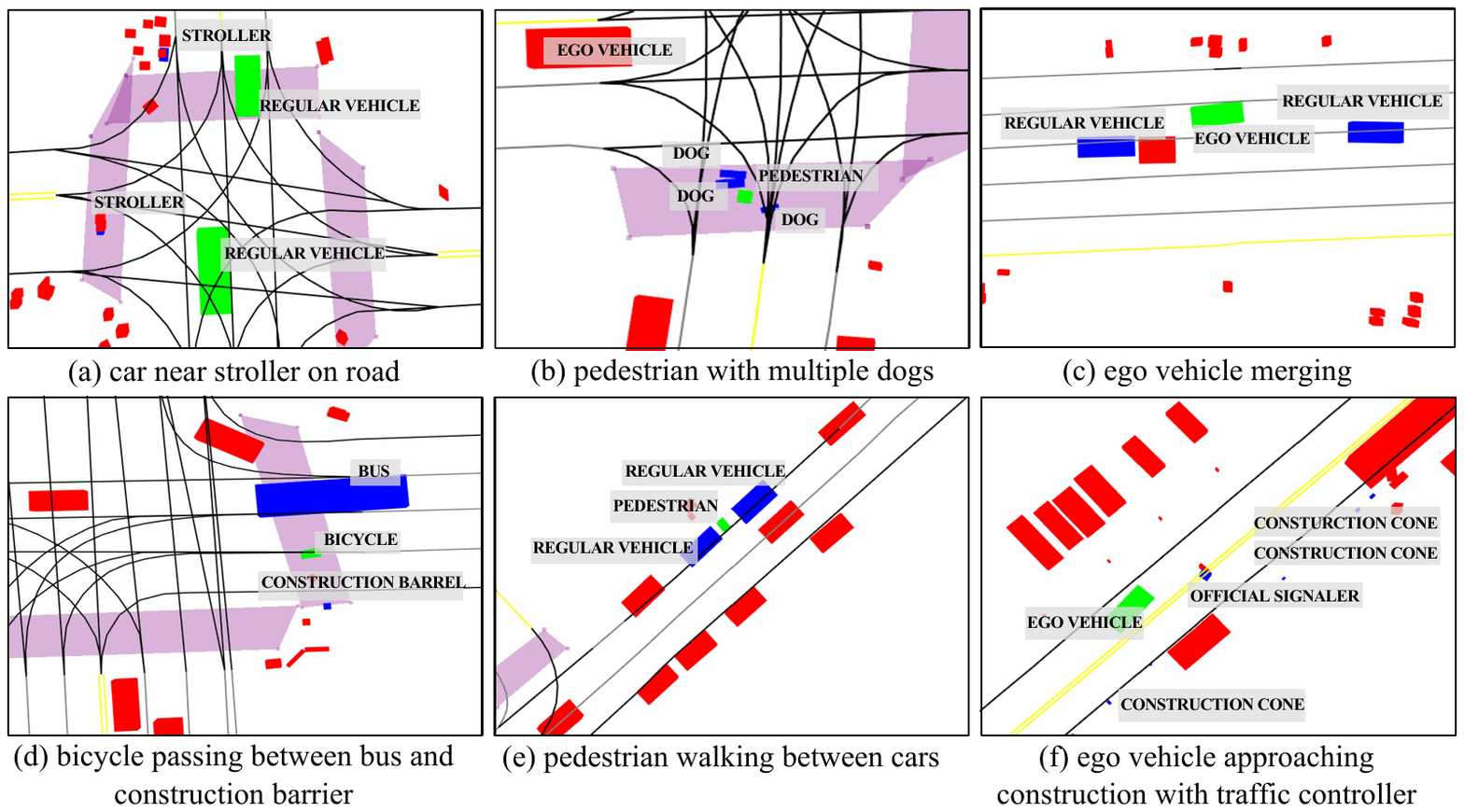}
    \caption{{\bf Examples of Multi-Agent Interactions.} We visualize  representative examples from RefAV to highlight the diversity of our dataset. In (a), we capture the interactions between vulnerable road users and vehicles at a crowded intersection. Scenario (b) presents an atypical instance of a common multi-agent interaction (e.g. {\tt pedestrian walking a dog}). In (c), we show a complex ego-vehicle trajectory that involves multiple moving vehicles. Scenario (d) illustrates an example of a rare multi-object interaction. In (e), we highlight a scenario that might require evasive maneuvers from the ego-vehicle (e.g. the occluded pedestrian might cross the path of the ego-vehicle). Finally, subfigure (f) visualizes a scenario with a multiple-step relationship (e.g. the official signaler is standing inside of a construction zone). Note that we show \textcolor{green!80!black}{\tt referred objects} in green, \textcolor{blue}{\tt related objects} in blue, and all \textcolor{red}{\tt other objects} in red. 
    }
    \label{fig:dataset_examples}
\end{figure*}

\textbf{Manual Scenario Annotation.} Despite the versatility of our procedural generation approach, many interesting interactions cannot be easily identified by composing atomic functions. Therefore, we manually inspect all videos in the validation and test sets to identify interesting driving behaviors (cf. Fig \ref{fig:dataset_examples}). Further, we manually annotate weather (e.g. {\tt clear}, {\tt cloudy}, {\tt rain}, {\tt snow}, and {\tt fog}) and lighting conditions (e.g. {\tt daylight}, {\tt dusk/dawn}, and {\tt night}) in the validation and test sets as these attributes are relevant for establishing a safety case \cite{lee2024typical} in all driving conditions. See the Appendix \ref{sec:annotation_tools} for more details about our annotation workflow.

{\bf Dataset Statistics.} As shown in Table \ref{tab:other_datasets}, our dataset uniquely addresses the task of spatio-temporal scenario mining, whereas other datasets focus on object detection, referring multi-object tracking (RMOT), or visual question answering (VQA). Further, RefAV is uniquely built on Argoverse 2 \cite{wilson2023argoverse}, whereas most existing datasets rely on nuScenes \cite{caesar2020nuscenes} or KITTI \cite{geiger2012we}. Next, RefAV provides track-level annotations at a higher frequency of 10Hz, compared to the typical 2Hz in other datasets, allowing for more fine-grained temporal analysis. While datasets like nuGrounding \cite{li2025nugrounding} and nuScenes-QA \cite{qian2024nuscenesqa} include large numbers of expressions, RefAV emphasizes a diversity of annotation types, including referring expressions that capture dynamic multi-agent interactions, weather, and lighting conditions. Lastly, RefAV is one of only two datasets to support negative prompts.
\begin{table}[b]
\caption{\textbf{Comparison to Other Benchmarks.} Language-based 3D scene understanding has been extensively studied in the context of referential multi-object tracking (RMOT) and multi-modal visual question answering (VQA). Different from prior work, we address the problem of spatio-temporal scenario mining. Specifically, RefAV is based on Argoverse 2, which provides 3D track-level annotations for 30 categories at 10 Hz. Although RefAV does not include as many referential expressions as prior work (e.g. OmniDrive \cite{wang2024omnidrive} and nuGrounding \cite{li2025nugrounding}), our referential annotations focus on capturing diverse multi-agent interactions. Lastly, RefAV includes negative prompts, which allows us to more accurately measure scenario mining performance. 
} 
\label{tab:other_datasets}
\resizebox{\linewidth}{!}{
\begin{tabular}{|l|c|c|c|c|c|c|c|c|c|}
\hline
\rowcolor{gray!10}
\multicolumn{1}{|c|}{\textbf{Dataset}} & \multicolumn{1}{c|}{\textbf{Base Data}} & \multicolumn{1}{c|}{\textbf{Task}} & \multicolumn{1}{c|}{\textbf{View}} & \multicolumn{1}{c|}{\textbf{Anno. Freq.}} & \multicolumn{1}{c|}{\textbf{\# Expressions}} & \multicolumn{1}{c|}{\textbf{\# Frames}} & \multicolumn{1}{c|}{\textbf{Neg. Prompts}} & \multicolumn{1}{c|}{\textbf{Human Anno.}} \\ \hline
Talk2Car \cite{deruyttere2019talk2car}                         &            nuScenes                        &        Detection                          &         Front                                  &        Per Video                               &       12k                                  &                   9k                      &                \textcolor{red}{\xmark}                          &             Referring Expression                                                                                                                                                                                    \\ \hline
nuScenes-QA   \cite{qian2024nuscenesqa}                         &         nuScenes                           &            VQA                      &        360$^\circ$                                    &       2Hz                                &             460k                            &           34k                               &              \textcolor{darkgreen}{\checkmark}                            &        None                                                                                                                                                                                        \\ \hline
DriveLM   \cite{sima2024drivelm}                         &       nuScenes                             &             VQA                     &               360$^\circ$                            &         2Hz                              &                443k                         &           5k                              &                     \textcolor{red}{\xmark}                     &                   QA Pairs                                                                                                                                                                              \\ \hline
OmniDrive   \cite{wang2024omnidrive}                         &          nuScenes                          &              VQA                    &            360$^\circ$                                &           2Hz                            &              200k                           &            34k                              &              \textcolor{red}{\xmark}                            &       None                                                                                                                                                                                          \\ \hline
Refer-KITTI  \cite{wu2023referkitti}                         &              KITTI                      &        RMOT                          &              Front                             &               2Hz                        &           818                              &               7k                          &                     \textcolor{red}{\xmark}                     &                   Referring Expression                                                                                                                                                                               \\ \hline
nuGrounding \cite{li2025nugrounding}                              &             nuScenes                       &         RMOT                         &       360$^\circ$                                    &      2Hz                                 &       2.2M                                  &          34k                               &                    \textcolor{red}{\xmark}                      &        Object Attribute                                                                                                                                                                                         \\ \hline
nuPrompt \cite{wu2023nuprompt}                              &          nuScenes                          &       RMOT                           &              360$^\circ$                              &         2Hz                              &       407k                                  &              34k                            &                    \textcolor{red}{\xmark}                      &         Object Color                                                                                                                                                                                       \\ \hline
RefAV                             &        Argoverse 2                            &          \begin{tabular}[c]{@{}c@{}}RMOT \&\\ Scenario Mining \end{tabular}                        &          360$^\circ$                                  &             10Hz                          &             10k                            &            155k                             &            \textcolor{darkgreen}{\checkmark}                              &          \begin{tabular}[c]{@{}c@{}}Referring Expression,\\ Weather, Lighting \end{tabular}                                                                                                                                                                                        \\ \hline
\end{tabular}
}
\end{table}

\subsection{Scenario Mining Baselines}
\label{ssec:baselines}

\begin{figure}[t]
    \centering
    \includegraphics[width=\linewidth,trim={0 3.7cm 0 0},clip]{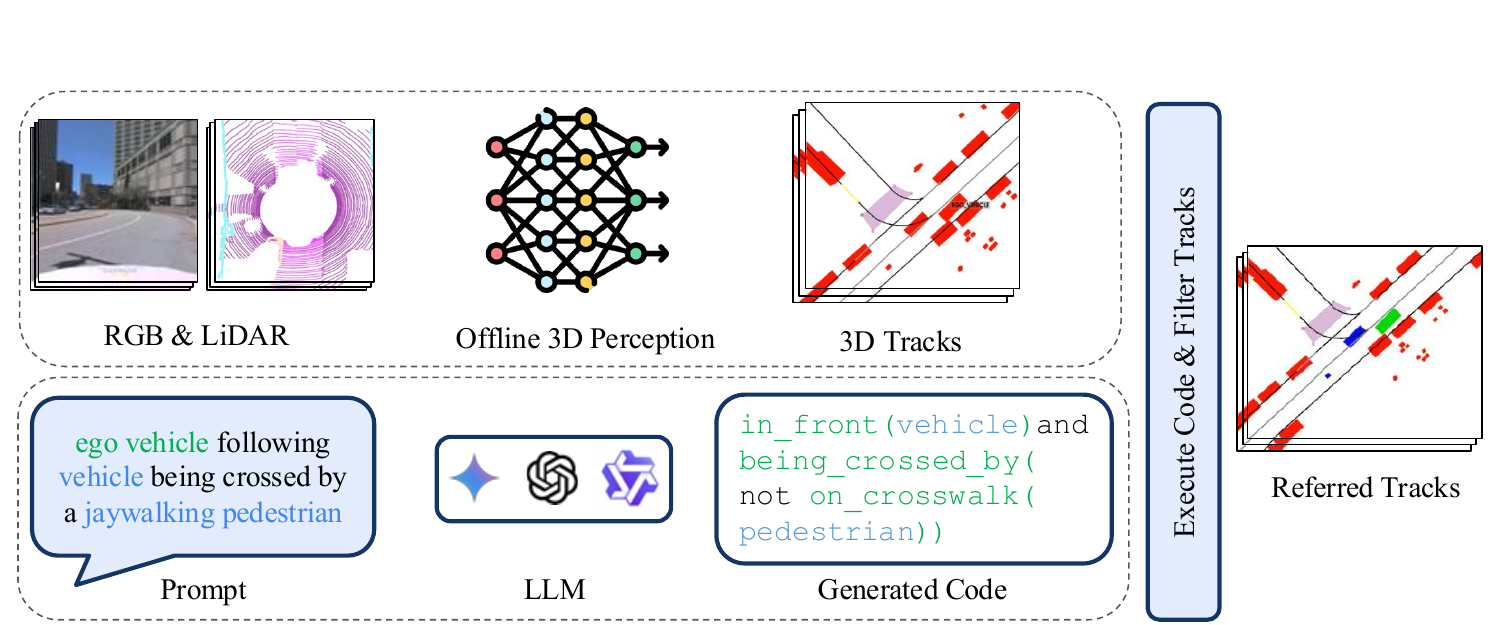}
    \caption{{\bf Method Overview.} RefProg is a dual-path method that independently generates 3D perception outputs and Python-based programs for referential grounding. Given raw LiDAR and RGB inputs, RefProg runs a offline 3D perception model to generate high quality 3D tracks. In parallel, it prompts an LLM to generate code to identify the referred track. Finally, the generated code is executed to filter the output of the offline 3D perception model to produce a final set of \textcolor{green!80!black}{\tt referred objects}, \textcolor{blue}{\tt related objects}, and \textcolor{red}{\tt other objects}.}
    \label{fig:baseline}
\end{figure}

We present five referential tracking baselines for scenario mining, including filtering by referred class, ReferGPT \cite{chamiti2025refergpt}, image embedding similarity, LLM APIs as a black box, and referential tracking by program synthesis. We repurpose 3D trackers from the \href{https://eval.ai/web/challenges/challenge-page/2006/overview}{AV2 End-to-End Forecasting Challenge}

\textbf{Filtering by Referred Class.} We propose a simple ``blind'' baseline that does not explicitly reason about multi-agent interactions. For a given natural language prompt, we use an LLM to parse the {\tt referred object} class and only keep 3D tracks from this class. For example, for the natural language query {\tt find all cars turning left}, we remove all predicted tracks except {\tt cars}. 


\textbf{ReferGPT.}  ReferGPT \cite{chamiti2025refergpt} classifies the output of off-the-shelf 3D trackers into {\tt referred objects} and {\tt other objects} using CLIP text embeddings. First, ReferGPT projects the predicted 3D bounding box of each track onto the 2D image plane at each timestamp to get an image crop of each object. Next, ReferGPT parses the object's coordinates, velocity, yaw, yaw rate, and distance from the ego-vehicle. Using both the object image crop and bounding box descriptor, ReferGPT uses a VLM to generate a descriptive caption. Finally, ReferGPT uses the text cosine-similarity between the generated caption and referential prompt to score the relevance of the object track. We use GPT-5-mini as our captioning VLM and SigLIP2 \cite{tschannen2025siglip2multilingualvisionlanguage} to compute text embeddings. We make several modifications to ReferGPT to adapt it for the spatio-temporal scenario mining problem. Since an object may be visible in multiple cameras at a single timestamp, we only caption the 2D bounding box with the largest area at each timestamp. ReferGPT selects {\tt referred objects} by clustering tracks with the highest similarity scores. Since RefAV contains also many negative prompts, we consider a track to be a {\tt referred object} if the similarity score between the caption text embeddings and prompt text embedding are above a fixed threshold for more than 50\% of the predicted track length.

\textbf{Image-Embedding Similarity.} We take inspiration from ReferGPT to classify the output of off-the-shelf 3D trackers into {\tt referred objects} and {\tt other objects} by directly using CLIP \cite{clip} image features. We extract the track image crops in the same way as ReferGPT and compute CLIP image features for each track at each timestep. Next, we compute the cosine similarity between the per-timestep CLIP image embeddings and the CLIP text embedding of the referential prompt. We consider a track to be a {\tt referred object} if the similarity score between the image and text embeddings are above a fixed threshold for more than 50\% of the track length.

\textbf{LLM API as a Black Box.} We evaluate GPT-5 as a black-box scenario mining algorithm. For each log in AV2, we ask the  OpenAI Responses API to identify which predictions match the referential prompt, and provide map data, city to ego-vehicle transforms, and 3D track information from an off-the-shelf tracker. GPT-5 responds with a CSV listing all {\tt referred objects} from the initial set of predictions. Interestingly, analyzing the thinking traces of GPT-5 shows that it writes code from scratch to parse the input files and reason about multi-agent interactions.

\textbf{Referential Tracking by Program Synthesis (RefProg).} RefProg takes a similar approach as LLM API as a Black Box, but provides more code scaffolding to improve accuracy. Specifically, we prompt an LLM to generate code using an API listing of atomic actions for a given referential query. We execute the generated code on off-the-shelf 3D tracks to identify {\tt referred objects}, {\tt related objects} and {\tt other objects} (cf. Fig \ref{fig:baseline}). In addition to using API listings for estimating an object's state and relationships from 3D tracks, RefProg is given access to visual tools such as SigLIPv2 to identify visual attributes of tracked objects. Despite the similarity of RefProg and our LLM-based procedural scenario generator, we highlight two key differences. First, RefProg generates code based on the referential query, while our LLM-based procedural scenario generator synthesizes the referential query based on the generated code. We posit that the role of the scenario generator is considerably easier since there are many valid natural language prompts to describe a generated program, but significantly fewer valid programs corresponding to each natural language prompt. Lastly, code from our scenario generator is extensively modified by human annotators for correctness, while code from RefProg is executed without verification. Notably, we also evaluate RefProg on nuPrompt to demonstrate that the atomic functions generalize beyond RefAV.

\section{Experiments}
\label{sec:experiments}

In this section, we briefly describe our evaluation metrics, provide an empirical analysis of our baselines, ablate the impact of different LLMs on RefProg's spatio-temporal scenario mining accuracy, and evaluate RefProg on nuPrompt to demonstrate the robustness of our approach.

\begin{table*}[t]
\tiny
\caption{\textbf{Experimental Results.} We evaluate several zero-shot referential tracking baselines. We find that RefProg significantly outperforms all other zero-shot baselines. Notably, filtering by referred class is a particularly strong baseline, outperforming image-embedding similarity. Interestingly, directly using LLM APIs as a black box outperforms ReferGPT's hand-crafted approach. All winning submissions to our challenge build upon RefProg. 
} 
\label{tab:experimental_results}
\resizebox{\linewidth}{!}{
\begin{tabular}{|lcccccc|}
\hline
\rowcolor{gray!10}
\multicolumn{1}{|l|}{\textbf{Method}} & \multicolumn{1}{l||}{\textbf{HOTA $\uparrow$}} & \multicolumn{1}{l|}{\textbf{HOTA-Temporal $\uparrow$}} & \multicolumn{1}{l|}{\textbf{HOTA-Track $\uparrow$}} & \multicolumn{1}{l|}{\textbf{Log Bal. Acc. $\uparrow$}} & \multicolumn{1}{l|}{\textbf{Timestamp Bal. Acc. $\uparrow$}}\\ \hline
\multicolumn{1}{|l|}{Ground Truth}         & \multicolumn{1}{c||}{100.0}              & \multicolumn{1}{c|}{13.3}        & \multicolumn{1}{c|}{20.5}            &      \multicolumn{1}{c|}{50.0}     &      \multicolumn{1}{c|}{50.0}           \\ \hline
\rowcolor{gray!10}
\multicolumn{6}{|l|}{\bf{Filter by Referred Class}}     \\ \hline
\multicolumn{1}{|l|}{Ground Truth}         & \multicolumn{1}{c||}{100.0}              & \multicolumn{1}{c|}{21.4}        & \multicolumn{1}{c|}{33.9}            &      \multicolumn{1}{c|}{52.9}     &      \multicolumn{1}{c|}{57.2}           \\ \hline
\multicolumn{1}{|l|}{LE3DE2E \cite{chenle3de2e}}         & \multicolumn{1}{c||}{74.4}              & \multicolumn{1}{c|}{19.2}        & \multicolumn{1}{c|}{30.0}            &      \multicolumn{1}{c|}{53.4}     &      \multicolumn{1}{c|}{55.0}           \\ \hline
\rowcolor{gray!10}
\multicolumn{6}{|l|}{\bf{Image-Embedding Similarity}}                                                                                                                                                    \\ \hline
\multicolumn{1}{|l|}{Ground Truth}         & \multicolumn{1}{c||}{100.0}              & \multicolumn{1}{c|}{24.6}        & \multicolumn{1}{c|}{28.3}            &      \multicolumn{1}{c|}{54.7}     &      \multicolumn{1}{c|}{57.0}           \\ \hline
\multicolumn{1}{|l|}{Le3DE2E \cite{chenle3de2e}}         & \multicolumn{1}{c||}{74.4}              & \multicolumn{1}{c|}{17.2}        & \multicolumn{1}{c|}{24.4}            &      \multicolumn{1}{c|}{51.1}     &      \multicolumn{1}{c|}{51.1}           \\ \hline
\multicolumn{1}{|l|}{ReVoxelDet \cite{lee2024revoxeldet}}         & \multicolumn{1}{c||}{63.1}              & \multicolumn{1}{c|}{17.1}        & \multicolumn{1}{c|}{21.2}            &      \multicolumn{1}{c|}{52.3}     &      \multicolumn{1}{c|}{54.4}           \\ \hline
\multicolumn{1}{|l|}{TransFusion \cite{bai2022transfusion}}         & \multicolumn{1}{c||}{63.6}              & \multicolumn{1}{c|}{16.0}        & \multicolumn{1}{c|}{18.2}            &      \multicolumn{1}{c|}{52.8}     &      \multicolumn{1}{c|}{55.3}           \\ \hline

\rowcolor{gray!10}
\multicolumn{6}{|l|}{\bf{ReferGPT \cite{chamiti2025refergpt}}}\\\hline
\multicolumn{1}{|l|}{Le3DE2E \cite{chenle3de2e}}         & \multicolumn{1}{c||}{74.4}              & \multicolumn{1}{c|}{20.2}        & \multicolumn{1}{c|}{30.8}&
\multicolumn{1}{c|}{57.0}     &      \multicolumn{1}{c|}{57.1}           \\ \hline

\rowcolor{gray!10}
\multicolumn{6}{|l|}{\bf{LLM APIs as a Black Box}}\\ \hline
\multicolumn{1}{|l|}{Le3DE2E \cite{chenle3de2e}}         & \multicolumn{1}{c||}{74.4}              & \multicolumn{1}{c|}{37.2}        & \multicolumn{1}{c|}{39.2}&
\multicolumn{1}{c|}{58.4}     &      \multicolumn{1}{c|}{62.3}           \\ \hline
\rowcolor{gray!10}
\multicolumn{6}{|l|}{\bf{Referential Tracking by Program Synthesis}}                                                                                                                                                    \\ \hline
\multicolumn{1}{|l|}{Ground Truth}         & \multicolumn{1}{c||}{100.0}              & \multicolumn{1}{c|}{64.8}        & \multicolumn{1}{c|}{68.7}            &      \multicolumn{1}{c|}{81.1}     &      \multicolumn{1}{c|}{80.7}           \\ \hline
\multicolumn{1}{|l|}{Le3DE2E \cite{chenle3de2e}}         & \multicolumn{1}{c||}{74.4}              & \multicolumn{1}{c|}{50.1}        & \multicolumn{1}{c|}{51.1}            &      \multicolumn{1}{c|}{\textbf{71.8}}     &      \multicolumn{1}{c|}{74.6}           \\ \hline
\multicolumn{1}{|l|}{ReVoxelDet \cite{lee2024revoxeldet}}         & \multicolumn{1}{c||}{63.1}              & \multicolumn{1}{c|}{43.9}        & \multicolumn{1}{c|}{45.1}            &      \multicolumn{1}{c|}{70.7}     &      \multicolumn{1}{c|}{69.8}           \\ \hline
\multicolumn{1}{|l|}{TransFusion \cite{bai2022transfusion}}         & \multicolumn{1}{c||}{63.6}              & \multicolumn{1}{c|}{48.1}        & \multicolumn{1}{c|}{46.6}            &      \multicolumn{1}{c|}{70.7}     &      \multicolumn{1}{c|}{68.7}           \\ \hline
\multicolumn{1}{|l|}{Valeo4Cast \cite{xu2024valeo4cast}}         & \multicolumn{1}{c||}{69.2}              & \multicolumn{1}{c|}{41.7}        & \multicolumn{1}{c|}{44.1}            &      \multicolumn{1}{c|}{62.7}     &      \multicolumn{1}{c|}{69.1}           \\ \hline
\rowcolor{gray!10}
\multicolumn{6}{|l|}{\bf{Challenge Submissions}}                                                                                                                                                    \\ \hline
\multicolumn{1}{|l|}{Zeekr UMCV}         & \multicolumn{1}{c||}{74.4}              & \multicolumn{1}{c|}{\textbf{53.4}}        & \multicolumn{1}{c|}{51.0}            &      \multicolumn{1}{c|}{66.3}     &      \multicolumn{1}{c|}{76.6}           \\ \hline
\multicolumn{1}{|l|}{Mi3 UCM}         & \multicolumn{1}{c||}{74.4}              & \multicolumn{1}{c|}{52.4}        & \multicolumn{1}{c|}{\textbf{51.5}}            &      \multicolumn{1}{c|}{65.8}     &      \multicolumn{1}{c|}{\textbf{77.5}}           \\ \hline
\multicolumn{1}{|l|}{ZXH}         & \multicolumn{1}{c||}{74.4}              & \multicolumn{1}{c|}{52.1}        & \multicolumn{1}{c|}{50.2}            &      \multicolumn{1}{c|}{66.5}     &      \multicolumn{1}{c|}{76.1}           \\ \hline
\end{tabular}
}
\end{table*}

\textbf{Metrics.} We evaluate scenario mining methods using HOTA-Temporal and HOTA-Track (variants of HOTA \cite{luiten2021hota}) to measure referential tracking performance and balanced accuracy to measure information retrieval accuracy. HOTA is a unified metric that explicitly balances detection accuracy, association, and localization, making it better aligned with human visual evaluations of tracking performance. HOTA-Temporal extends the standard HOTA metric by only considering the referred timestamps within a track as true positives. HOTA-Track is similar to HOTA-Temporal, but it does not penalize methods for incorrectly predicting the start and end of a referred action. For example, given the prompt {\tt car turning right}, HOTA-Temporal only considers the timestamps where the car is turning right in the full referred track as true positives, whereas HOTA-Track considers all timestamps in the referred track as true positives. We also include HOTA results to contextualize standard tracking performance with the referential tracking performance of all benchmarked trackers. Additionally, we use timestamp balanced accuracy as a timestamp-level classification metric and log balanced accuracy as a log-level classification metric to evaluate how well methods can identify which timestamps and logs contain objects that match the referential prompt, respectively. Balanced accuracy is used over binary classification metrics such as F1 due to the imbalance of positive and negative prompts.

For each prompt, we categorize all objects into three groups: {\tt referred objects}, which are the primary objects specified by the prompt; {\tt related objects}, which are objects that interact with the referred object; and {\tt other objects}, which are neither referred to nor relevant to the prompt. We compute both HOTA-Temporal and HOTA-Track metrics exclusively for the {\tt referred object} class in the main paper. We report the performance on {\tt related objects} and {\tt other objects} in the Appendix \ref{sec:related_other}.

\subsection{Empirical Analysis of Results}
\label{sec:results}

Table \ref{tab:experimental_results} presents several simple baselines for addressing spatio-temporal scenario mining. We evaluate each of the locally run zero-shot baselines using oracle ground-truth track annotations to understand how tracking accuracy and language grounding independently influence final performance. We do not evaluate {\it ReferGPT} and {\it LLM APIs as a Black Box} with the ground-truth track annotations to avoid dataset leakage. Evaluating every track as {\tt referred object} without considering language prompts (first row) yields a HOTA-Temporal of 13.3\% and HOTA-Track of 20.5\%. Predicting the nearest object class that corresponds to the prompt with {\it Filter by Referred Class} improves referential tracking accuracy by 8\%. However, {\it Filter by Referred Class} with LE3DE2E tracks results in a slight performance drop over ground truth tracks. This performance drop is perhaps lower than expected because Le3DE2E has exceptionally high tracking accuracy for {\tt REGULAR VEHICLE}, which dominates the AV2 dataset. The {\it Image-Embedding Similarity} oracle outperforms the {\it Filter by Referred Class} oracle, suggesting that CLIP-based filtering can capture richer semantics than class names alone. However, {\it Image-Embedding Similarity} underperforms {\it Filter by Referred Class} when using predicted tracks (e.g., LE3DE2E scores 17.2 vs. 19.2 HOTA-Temporal). This likely stems from CLIP being much more sensitive to the predicted 3D location of the bounding box. {\it ReferGPT} slightly outperforms {\it Filter by Referred Class} and {\it Image-Embedding Similarity} across all reported metrics. {\it ReferGPT}, unlike the other methods, uses information about the current position and velocity of an object relative to the ego vehicle. Notably, the {\it LLMs as a Black Box} baseline significantly improves over prior approaches with a HOTA-Temporal of 37.2\%. Finally, our proposed {\it RefProg} baseline outperforms all other zero-shot methods. Notably, RefProg with LE3DE2E tracks achieves a 13.8\% improvement in HOTA-Temporal over {\it LLMs as a Black Box} with Le3DE2E tracks.  

\textbf{Scenario Mining Challenge.} 
We hosted a \href{https://cvpr2025.wad.vision/}{challenge at CVPR 2025} to encourage broad community involvement in addressing spatio-temporal scenario mining. Our competition received submissions from eight teams. Notably, four teams beat our best baseline. We present the top three entries at the bottom of Table \ref{tab:experimental_results}, summarize their contributions in Appendix \ref{sec:cvpr_results} and include a link to full technical reports and code \href{https://www.neeharperi.com/WAD25-Scenario-Mining-Challenge}{here}. The best performing team beats our baseline by 3.3\% achieving 53.4 HOTA-Temporal.

\begin{table}[t]
\caption{\textbf{Impact of Different LLMs on Code Synthesis Quality.} We evaluate the impact of using different LLMs for program synthesis in the RefProg pipeline. Interestingly, we find that Claude 3.5 Sonnet has the lowest program failure rate (e.g. 99.5\% of Claude 3.5 Sonnet's generated programs were valid, compared to 81.9\% for Qwen 2.5B Instruct), while Claude 3.7 Sonnet achieves the highest HOTA-Temporal. 
} 
\label{tab:llm_quality}
\resizebox{\linewidth}{!}{
\begin{tabular}{|lcccccc|}
\hline
\rowcolor{gray!10}
\multicolumn{1}{|l|}{\textbf{LLM}} & \multicolumn{1}{l|}{\textbf{Failure Rate $\downarrow$}} & \multicolumn{1}{l|}{\textbf{HOTA-Temporal $\uparrow$}} & \multicolumn{1}{l|}{\textbf{HOTA-Track $\uparrow$}} & \multicolumn{1}{l|}{\textbf{Log Bal. Acc. $\uparrow$}} & \multicolumn{1}{l|}{\textbf{Timestamp Bal. Acc. $\uparrow$}}\\ \hline
\multicolumn{1}{|l|}{Qwen-2.5-7B-Instruct}         & \multicolumn{1}{c|}{18.1}              & \multicolumn{1}{c|}{31.6}        & \multicolumn{1}{c|}{34.4}            &      \multicolumn{1}{c|}{62.1}     &      \multicolumn{1}{c|}{62.0}           \\ \hline
\multicolumn{1}{|l|}{gemini-2.0-flash}         & \multicolumn{1}{c|}{2.6}              & \multicolumn{1}{c|}{45.2}        & \multicolumn{1}{c|}{46.6}            &      \multicolumn{1}{c|}{{\bf 72.1}}     &      \multicolumn{1}{c|}{{\bf 74.6}}           \\ \hline
\multicolumn{1}{|l|}{gemini-2.5-flash-preview-04-17}         & \multicolumn{1}{c|}{15.4}              & \multicolumn{1}{c|}{47.8}        & \multicolumn{1}{c|}{47.6}            &      \multicolumn{1}{c|}{71.0}     &      \multicolumn{1}{c|}{73.8}           \\ \hline
\multicolumn{1}{|l|}{claude-3.5-sonnet-20241022}         & \multicolumn{1}{c|}{{\bf 0.5}}              & \multicolumn{1}{c|}{46.1}        & \multicolumn{1}{c|}{47.5}            &      \multicolumn{1}{c|}{71.8}     &      \multicolumn{1}{c|}{71.8}           \\ \hline
\multicolumn{1}{|l|}{claude-3.7-sonnet-20250219}         & \multicolumn{1}{c|}{2.9}              & \multicolumn{1}{c|}{{\bf 50.1}}        & \multicolumn{1}{c|}{{\bf 51.1}}            &      \multicolumn{1}{c|}{71.8}     &      \multicolumn{1}{c|}{{\bf 74.6}}           \\ \hline
\end{tabular}}
\end{table}



\textbf{Impact of LLMs on Code Generation.} We evaluate the impact of using different LLMs on code generation quality in RefProg. We use Le3DE2E's tracks and a consistent prompt for all experiments in this ablation. We find that Claude 3.7 Sonnet performs the best (cf. Table \ref{tab:llm_quality}), achieving a HOTA-Temporal of 50.1. We posit that this is because it has been explicitly tuned for code generation and instruction following. Gemini 2.0 Flash performs considerably worse. Interestingly, we find that Gemini-2.5-Flash and Qwen-2.5-Instruct struggle to generate valid programs, and instead try to generate full programs using invalid import statements. Failure Rate indicates the percentage of generated programs that throw an exception while running.

\textbf{Evaluating RefProg on nuPrompt.} The atomic functions used in RefAV's dataset generation and RefProg represent fundamental motion primitives that are generalizable across datasets. To this end, we evaluate RefProg on nuPrompt \cite{wu2023nuprompt}, a popular referential-tracking dataset based on nuScenes in Table \ref{tab:nuprompt}. Notably, we find that our zero-shot compositional approach outperforms current state-of-the-art methods trained on domain-specific data. We make minimal adaptations to RefProg for the nuScenes dataset. We swap out the AV2 annotated class list for the nuScenes annotated class list and remove atomic functions that require access to an HD map. Importantly, we make zero modifications to our atomic action definitions, and expect that  dataset-specific modifications can improve performance further. 

\begin{table}[t]
\centering
\caption{\textbf{Zero-Shot Evaluation on nuPrompt.} We evaluate RefProg on nuPrompt and achieve state-of-the-art accuracy, highlighting the strong generalization of our approach. Importantly, we do not modify RefProg's atomic action definitions.}
\label{tab:nuprompt}
\resizebox{\linewidth}{!}{
\begin{tabular}{|l|c|c|c|c|c|}
\hline
\rowcolor{gray!10}
\textbf{Method}                     & \textbf{Decoder} & \textbf{AMOTA} $\uparrow$ & \textbf{AMOTP} $\downarrow$ & \textbf{Recall} $\uparrow$ & \textbf{MOTA} $\uparrow$  \\ \hline
CenterPoint Tracker \cite{yin2021center}                 & DETR3D           & 0.079          & 1.820          & 0.196           & 0.093          \\ \hline
CenterPoint Tracker \cite{yin2021center}                & PETR             & 0.178          & 1.650          & 0.291           & 0.197          \\ \hline
DQTrack \cite{li2023dqtrack}                            & DETR3D           & 0.186          & 1.641          & 0.307           & 0.208          \\ \hline
DQTrack  \cite{li2023dqtrack}                           & Stereo           & 0.198          & 1.625          & 0.309           & 0.214          \\ \hline
DQTrack  \cite{li2023dqtrack}                           & PETRv2           & 0.234          & 1.545          & 0.332           & 0.269          \\ \hline
ADA-Track \cite{ding2024ada}                           & PETR             & 0.249          & 1.538          & 0.353           & 0.270          \\ \hline
PromptTrack \cite{wu2023nuprompt}                        & DETR3D           & 0.202          & 1.615          & 0.310           & 0.222          \\ \hline
PromptTrack  \cite{wu2023nuprompt}                      & PETR             & 0.259          & 1.513          & 0.366           & 0.280          \\ \hline
RefProg (Ours) & StreamPETR \cite{wang2023exploring}                & \textbf{0.321} & \textbf{1.238} & \textbf{0.504}  & \textbf{0.329} \\ \hline
\end{tabular}
}
\end{table}


\textbf{Analysis of Failure Cases.} While RefProg outperforms \textit{Image-Embedding Similarity} and \textit{ReferGPT}, each method has unique failure cases. We find that \textit{Image-Embedding Similarity} performs poorly because the CLIP image embedding for each track lacks the context required to understand multi-agent interactions. Specifically, the 2D projected image only includes the tracked object, and does not contain information about past or future frames. \textit{ReferGPT} similarly suffers from a lack of temporal context. Lastly, we find that RefProg fails in cases where the API listing is not expressive enough for a given prompt (e.g., prompts involving weather and lighting conditions). Although one can always add new atomic actions, this strategy is not scalable. 

\textbf{Limitations and Future Work.} The quality of our scenario mining dataset is limited by the quality of the ground truth 3D perception labels in AV2. For example, jittery tracks can lead to poor motion classification over short horizons. We address this issue by significantly post-processing and manually verifying all generated scenarios to mitigate label noise.  Although AV2 includes many interesting scenarios (cf. Fig. \ref{fig:dataset_examples}), it is still relatively small compared to industry-scale datasets. Future datasets should be explicitly curated to address spatiotemporal scenario mining.

\section{Conclusion}
\label{sec:conclusion}
In this paper, we introduce RefAV, a large-scale benchmark designed to evaluate scenario mining. Unlike prior language-based 3D scene understanding tasks, we find that scenario mining poses unique challenges in identifying complex multi-agent interactions. Notably prior referential tracking baselines struggle on this challenging benchmark, demonstrating the limitations of existing methods. Future work should develop models capable of reasoning over complex, multi-modal temporal data.


\newpage

{
    \small
    \bibliographystyle{ieeenat_fullname}
    \bibliography{main}
}

\clearpage
\newpage
\appendix

\section{Implementation Details}
\label{sec:impl_details}

We present additional implementation details to reproduce our baseline experiments below. We primarily consider two-stage baselines that first track all objects and then classify them into {\tt referred objects}, {\tt related objects}, or {\tt other objects}. We repurpose 3D trackers from winning teams of the Argoverse 2 End-to-End Forecasting Challenge. All models are trained to output tracks for 26 annotated categories within Argoverse 2. We include an example prediction signature below: 
\begin{lstlisting}[style=pythonstyle,language=Python]
    {
          "timestamp_ns": int,
          "track_id": int,
          "confidence": float,
          "class_name": str,
          "translation_m": np.array,
          "size": np.array,
          "yaw": np.array,
    }
\end{lstlisting}

We append a track for the ego vehicle with a consistent \texttt{track\_id}, confidence of 1.00, class name of \texttt{EGO\_VEHICLE}, yaw of 0 radians, size (<length, width height> in meters) of <4.877, 2, 1.473>, and translation (<x, y, z> in meters) of <1.422, 0, 0.25>. This translation represents the offset from the centroid of the ego-vehicle to the ego-vehicle reference coordinate  near the rear axle.

\textbf{All Tracks Oracle.} Our naive baseline labels all ground truth tracks as {\tt referred objects}. This means that all logs and timestamps are labeled as ``positive'', giving a balanced accuracy of 50\%. 

\textbf{Filter by Referred Class.}
This baseline filters all tracks except those corresponding to the referred class. To get the referred class, we prompt Claude 3.7 Sonnet as follows:

\greybox{
Categories: \{AV2 Categories\} 

Please select the one category that most corresponds to the object of focus in the description:\{Natural Language Description\}. As an example,
for the description ``vehicle turning at intersection with nearby pedestrians''
your output would be VEHICLE. For the description ``ego vehicle near construction barrel'' your output would be EGO\_VEHICLE. Your only output should
be the category name.}

We define two super-categories for \texttt{VEHICLE} and \texttt{ANY\_OBJECT}. The \texttt{VEHICLE} category is manually defined to contain the classes \texttt{ARTICULATED\_BUS, BOX\_TRUCK, BUS, EGO\_VEHICLE, LARGE\_VEHICLE, MOTORCYCLE, RAILED\_VEHICLE, REGULAR\_VEHICLE, SCHOOL BUS, TRUCK, and TRUCK\_CAB}.

\textbf{Image Embedding Similarity.}
This baseline filters tracks by computing prompt-track CLIP similarity over time. At each timestep, we project the eight vertices of each object's bounding box onto the image plane of all seven cameras. We then cull any objects that do not partially fall within the frustum of each camera. For each projected bounding box, if three or more vertices fall within the image plane, we take a crop using the minimum and maximum uv image coordinates. We pad this crop by 30 pixels on each side and feed the resulting image into a CLIP ViT-L/14 model to obtain CLIP image embeddings. We also use CLIP ViT-L/14 to compute text embeddings for all prompts in the dataset. Next, we calculate cosine similarity scores between each track's image crops and the text prompt. Since individual similarity scores tend to be noisy, we follow ReferGPT \cite{chamiti2025refergpt} in applying a dynamic threshold and majority voting strategy to identify the {\tt referred object} tracks. Since the dataset includes both positive and negative examples for each prompt, we group similarity scores by prompt and threshold the top 10\% of scores. If more than half of the scores of a track fall within the top 10\% threshold, we label the entire track as a {\tt referred object}. To reduce the number tracks of length one (i.e., tracks spanning only one timestamp), we apply a similarity score modifier: $s_{\text{modified}}=s_{\text{cosine}}-\frac{0.05}{len(\text{track})}$.

\textbf{LLM APIs as a Black Box} We use the \href{https://platform.openai.com/docs/api-reference/responses}{OpenAI Responses API} to determine whether particular scenarios occur within a data log. We prompt GPT-5 with three files: an HD map data, 6-DoF ego-to-city transformations, and object tracker predictions. We use gpt-5-2025-08-07 with medium reasoning effort and access to a code interpreter tool. We give instructions to save a CSV containing the {\tt referred object} id's and timestamps for each scenario. All 150 test data logs are run in parallel, with one request and one response for each data log. The saved CSVs are parsed into the RefAV format for evaluation as each response is delivered. We include the instructions given to GPT-5 below. 

\greybox{You are an expert data analyst. Your job is to identify if a given scenarios occurs within an autonomous driving log. 
A log contains files with an HD map, 6-DoF poses of the ego vehicle, and bounding box annotations from an object tracker. 

For each of the given scenario descriptions, output a CSV file containing the track\_uuids and timestamps of the referred objects. Some logs may not contain any referred objects. 
Your csv file should have ``track\_uuid'' and ``timestamp\_ns'' columns.
Your output csv files should have the naming convention scenario\{i\}.csv, depending on the order the descriptions are given. Make sure to save all of your CSVs! Identify if each of these scenarios occurs within the log.\\\{scenario\_descriptions\}
}

\textbf{ReferGPT.} This baseline uses a VLM to caption the cropped images of each track at each timestep.  Tracks are filtered by computing a text similarity score between the prompt and track captions. We take several steps to adapt ReferGPT to the RefAV dataset. At each timestep, we project the eight vertices of each predicted track's bounding box onto the image plane of all seven cameras. We then cull any objects that do not partially fall within the frustum of each camera. For each projected bounding box, if three or more vertices fall within the image plane, we take a crop using the minimum and maximum uv image coordinates.
If the predicted track is visible in multiple cameras, we use the 2D bounding box with the highest area. We pad this crop by 10$\%$ of the crop's length on each side. For the ego-vehicle, the crop is the entire image from the front camera. Each of the per object-timestamp crops is also paired with the motion statistics. We compute a track $i$'s motion statistics at timestep $t$ as $\mathbf{C}^i_t=\left[x,y,\theta,v_x,v_y,\alpha,d\right]$, where x and y are the position of the track in meters, $\theta$ is the yaw of the track in degrees, $v_x$ and $v_y$ are the velocities of the track in meters per second, $\alpha$ is the yaw rate of the track in degrees per second. $d$ is the distance from the centroids of the ego-vehicle and track in meters. All values are with respect to a static ego-centric coordinate frame. We prompt gpt-5-mini to caption each of the images as follows:
\greybox{
Caption this image taken from an autonomous vehicle (referred to as the ego-vehicle). Our detection model classifies this object as \{object\_category\}. 

Here are the coordinates of the object \{object\_motion\}. This is in the format [x, y, yaw, x\_velocity, y\_velocity, angular\_velocity, distance].

This is in a coordinate system where positive x is to the front of the ego vehicle, positive y is to the left of the ego-vehicle, and positive z is above the ego-vehicle. Position and velocity are measured in meters and meters per second. Yaw is measured in degrees counter-clockwise (colloquially ``left'') from the x-axis. Angular velocity is in degrees per second. Distance is distance from the ego-vehicle. 

Please describe the appearance of the object and scene, if they are located relative to the road/ego-vehicle, how they are moving, and how they are interacting with any nearby objects (if any are visible). 
Treat values of x, y, x\_velocity, and y\_velocity near zero as zero.
Feel free to add more relevant information in the captions.}

We present an example caption below:
\greybox{A motorcyclist slightly to the front-right of the ego-vehicle is stopped or creeping forward beside a compact SUV, straddling a scooter with feet down and hands on the handlebars. The rider is close to the SUV’s rear quarter, appearing stationary or moving very slowly while sharing the lane alongside other stopped traffic. The scene looks like an urban intersection or congested street with vehicles side-by-side and the rider positioned between them.}

To determine if an object is {\tt referred}, we compute a score $s = s_{cosine} + 0.1s_{fuzzy}$. $s_{cosine}$ is the cosine similarity between siglip2-base-patch16-naflex text embeddings of the caption and prompt. We follow the original ReferGPT implementation and compute $s_{fuzzy}$ using Gestalt pattern matching to find the similarity between two strings. Finally, we determine that an entire track is {\tt referred} if the majority the track's caption scores are above $0.63$. 

We find that the SigLIPv2 text similarity does not effectively separate the {\tt referred} and {\tt related objects} in the prompt. For example,  the query ``car to the left of a pedestrian'', has high similarity with both ``the vehicle is to the left of the person'' and ``the person is to the right of the vehicle''. To resolve this issue, we post-process predictions with \emph{Filter by Referred Class} to obtain the final set of {\tt referred objects}.

\textbf{Referential Tracking by Program Synthesis.}
RefProg filters off-the-shelf tracks using Python programs synthesized by a large language model (LLM), without explicitly relying on any information from LiDAR or cameras. To reduce the  time required to run these synthesized programs, we first filter the input tracks based on confidence. We compute each track's confidence score by summing its confidence values across all timestamps. For the classes {\tt REGULAR\_VEHICLE}, {\tt PEDESTRIAN}, {\tt BOLLARD}, {\tt CONSTRUCTION\_CONE}, and {\tt CONSTRUCTION\_BARREL}, we retain the top 200 tracks per class. For all other classes, we retain the top 100 tracks per category. We prompt the LLM as follows:

\greybox{Please use the following functions to find instances of a referred object in an autonomous driving dataset. Be precise to the description, try to avoid returning false positives. 

API Listing: \{API Listing\}

Categories: \{AV2 Categories\}

Define a single scenario for the description:\{Natural Language Description\}

Here is a list of examples: \{Prediction Examples\}. Only output code and comments as part of a Python block. Do not define any additional functions, or filepaths. Do not include imports. Assume the log\_dir and output\_dir variables are given. Wrap all code in one python block and do not provide alternatives. Output code even if the given functions are not expressive enough to find the scenario.
}

We include the API listing in Appendix \ref{sec:api_listing}. We found this prompt consistently produces reasonable code across all tested LLMs. After running the synthesized programs, we post-process the output tracks. We discard object relationships where the spatial distance exceeds 50 meters. For the remaining tracks, we apply timestamp dilation to reduce flickering of the \texttt{referred} object labels. We symmetrically dilate time segments to a minimum of 1.5 seconds. For example, if the base tracker tracks an object with ID 1 from 3.0 to 15.0 seconds at 2 Hz, and the synthesized code marks it as referred at timestamps 4.5 and 5.0, we update the final output to label the track as referred from 4.0 to 5.5 seconds. Finally, we downsample all output tracks to 2 Hz for evaluation.

\section{Token Statistics for LLM Baselines}
\textit{ReferGPT}, \textit{LLM APIs as a Black Box}, and \textit{RefProg} can be viewed as different ways of providing scaffolding around an LLM for RMOT and scenario mining tasks. Most commercially available LLM APIs bill according to four cost categories: uncached input tokens, cached input tokens, output tokens, and code interpreter sessions (CIS). Input token count scales approximately linearly with the length of the input text, the number of input images, and seconds of input video. \textit{ReferGPT} requires orders of magnitude more input and output tokens because it calls a VLM call for every tracked object at every timestamp. It also requires processing images, which can use more than 1000 tokens (depending on resolution) using GPT-5-mini. \textit{LLM APIs as a Black Box} only makes as many calls as are there logs of data. However, inspecting the reasoning traces reveals that GPT-5 automatically went through multiple rounds of generating data processing code and verifying the output. The LLM also often re-generates the same processing code between logs. Initializing code interpreter sessions represents approximately half of the total cost (\$$38.49$) of inference. RefProg makes as many calls as there are unique scenario descriptions. Since the entire atomic function API must be given in context, it requires a relatively high number of input tokens. However, most of these tokens are cached between requests. The output of the RefProg is relatively compact because the LLM has access to high-level composable functions and does not need to write extensive Python code. 

\begin{table}[h]
\caption{\textbf{Token Efficiency by LLM Baseline} RefProg achieves the highest scenario mining performance for the lowest cost. Cost is in USD for the OpenAI Platform in November 2025. We report the total number of tokens used millions. CIS refers to code interpreter sessions. }
\resizebox{\linewidth}{!}{
\begin{tabular}{|l|c|c|c|c|c|c|c|}
\hline
\rowcolor{gray!10}
\textbf{Method}      & \textbf{LLM} & \textbf{HOTA-Temporal $\uparrow$} & \textbf{Cost $\downarrow$} & \textbf{Uncached Input} & \textbf{Cached Input} & \textbf{Output} & \textbf{CIS} \\ \hline
ReferGPT             & GPT-5-mini   & 20.7                   & 150.64        & 372.9M                  & 4.0M                  & 28.7M           & 0            \\ \hline
LLM API as Black Box & GPT-5        & 37.2                   & 77.59         & 4.5M                    & 0                     & 3.4M            & 1283         \\ \hline
RefProg (Ours)             & GPT-5        & 42.3                   & 13.89         & 0.6M                    & 2.6M                  & 1.3M            & 0            \\ \hline
\end{tabular}}
\end{table}

\section{Comparison of Recent Datasets}

We provide a qualitative comparison of language expressions found in RefAV with other popular RMOT-for-driving and VQA-for-driving datasets. In particular, we examine nuPrompt, Refer-KITTI, and nuScenes-QA. To obtain a representative sample of each dataset, we compute text embeddings for each expression using Qwen-3-Embedding-8B. We display the first 30 samples using farthest point sampling initialized from the two samples furthest apart from each other. Many expressions in nuPrompt deal with object attributes that are minimally relevant to driving such as car color, a person's gender, or a person's clothing. Refer-KITTI similarly only refers to annotated vehicle and pedestrian classes. nuScenes-QA asks questions that are relevant to driving, but the dataset is limited by its lack of human annotations and unnatural sounding questions. We present examples from each dataset below.

\lstdefinestyle{scenestyle}{
    basicstyle=\ttfamily\small,
    frame=single,
    breaklines=true,
    columns=flexible,
    keepspaces=true,
    showstringspaces=false
}

\begin{center}
\textbf{RefAV (Ours)} 
\end{center}
\begin{lstlisting}[style=scenestyle]
- the vehicle two cars behind the ego vehicle
- pedestrian walking with dog on sidewalk
- stationary object
- motorcycle at stop sign
- group of people
- construction workers working next to the road
- fire truck in a non-emergency
- excavator
- large truck blocking view of vehicle that is about to turn left onto a main road
- bicyclist changing lanes to the right
- car with a stroller to its left
- wheelchair user at pedestrian crossing
- person directing traffic in a school zone
- skateboarder on the road
- unattended dog
- accelerating wheeled devices
- construction barrel with at least 2 construction cones within 3 meters
- bus with multiple pedestrians waiting on the right side
- vehicle passing the ego vehicle going signifcantly over the speed limit on an urban road
- van facing the wrong way in the middle turn lane
- two of the same work truck
- pedestrian near message board trailer
- pedestrians over 30 meters away
- the vehicle behind another vehicle being crossed by a jaywalking pedestrian
- vehicle flashing their hazard lights
- ego vehicle driving while it is cloudy
- motorcycle or pedestrian within 5 meters to the right of the ego vehicle
- bicycle behind a bus
- scooter lying down on the side of the road
- vehicles being passed by a motorcycle on either side
\end{lstlisting}

\newpage 

\begin{center}
\textbf{nuPrompt} \cite{wu2023nuprompt}
\end{center}

\begin{lstlisting}[style=scenestyle]
- Back, the pedestrian
- a stationary orange parking is on the parking space
- at the parking space, the trailer
- Car waiting with the same direction
- A person on foot by the roadside
- The silver lightcolored parked car is stationary in the parking space
- The objects possess the color blue
- Males with a black bag are getting ready to cross the road
- The one riding the black lefthandturn bicycle is a person
- Men stand on the street wearing pants and hats
- Females walking across the street
- Men wear pants while walking across the street
- In a state of sitting
- The car is in the act of turning
- Pedestrian getting ready to cross over
- is in the process of making a turn
- The car in motion is black and moving in the opposite direction
- Males wear tshirts and pants while making a left turn on the opposite side of the road
- the pedestrian is situated at the back left
- The dark car is at thestandstill and awaiting with the opposite direction as the truck
- The truck is thepale silver color
- we see the car ahead
- Females walk while wearing pants in the same direction
- Security camera
- Men are wearing shorts while standing on the road
- There is the truck positioned in the parking spot
- Color is red
- The bus is in front
- On the pedestrian pathway
- The car that is moving in the same direction is blue
\end{lstlisting}

\begin{center}
\textbf{Refer-KITTI} \cite{wu2023referkitti}
\end{center}

\begin{lstlisting}[style=scenestyle]
- cars which are faster than ours
- walking pedestrian in the left
- women back to the camera
- males in the right
- left cars which are parking
- people in the pants
- people
- women carrying a bag
- light-color vehicles
- vehicles in the counter direction
- walking males
- standing females
- men in the left
- right cars in silver
- red moving cars
- women
- walking people in the right
- vehicles
- pedestrian
- vehicles in front of ours
- vehicles which are in the left and turning
- vehicles in white
- cars in horizon direction
- black vehicles in right
- standing men
- vehicles which are braking
- walking women
- men
- right cars in the same direction of ours
- black cars
\end{lstlisting}

\begin{center}
\textbf{nuScenes-QA} \cite{qian2024nuscenesqa}
\end{center}

\begin{lstlisting}[style=scenestyle]
- How many without rider motorcycles are there?
- There is a thing that is both to the front of me and the front left of the traffic cone; what is it?
- Does the trailer have the same status as the truck that is to the back of the trailer?
- How many pedestrians are to the back of the moving bus?
- What is the with rider thing that is to the back right of the with rider bicycle?
- Are there any things?
- Is there another barrier of the same status as the barrier?
- The car to the back right of the stopped construction vehicle is in what status?
- There is a parked truck; what number of barriers are to the front left of it?
- Are there any other bicycles that in the same status as the car?
- What status is the motorcycle that is to the front left of the standing pedestrian?
- The without rider thing that is to the front of the construction vehicle is what?
- What number of traffic cones are there?
- Are there any trailers to the front of me?
- What is the thing to the back of the stopped thing?
- There is a without rider thing; are there any parked cars to the back of it?
- What number of barriers are to the back of the motorcycle?
- Are there any animals to the back right of me?
- The thing that is to the back right of the truck and the back right of the bus is what?
- There is a bus; is its status the same as the thing that is to the front of the bus?
- There is a with rider bicycle; what number of moving things are to the front of it?
- Are there any moving construction vehicles?
- The standing pedestrian to the back right of the traffic cone is what?
- How many buss are there?
- Is the status of the truck to the back of the pedestrian the same as the truck to the front of the bicycle?
- Is there another thing that has the same status as the animal?
- Are there any things to the back of the moving trailer?
- Are there any moving buss to the front of the without rider thing?
- The thing that is to the front of the moving truck and the front left of the stopped trailer is what?
- How many stopped trailers are there?
\end{lstlisting}

\section{Scenario Mining Annotation Tool}
\label{sec:annotation_tools}

\begin{figure}[H]
    \centering
    {
        \fbox{\includegraphics[width=\linewidth]{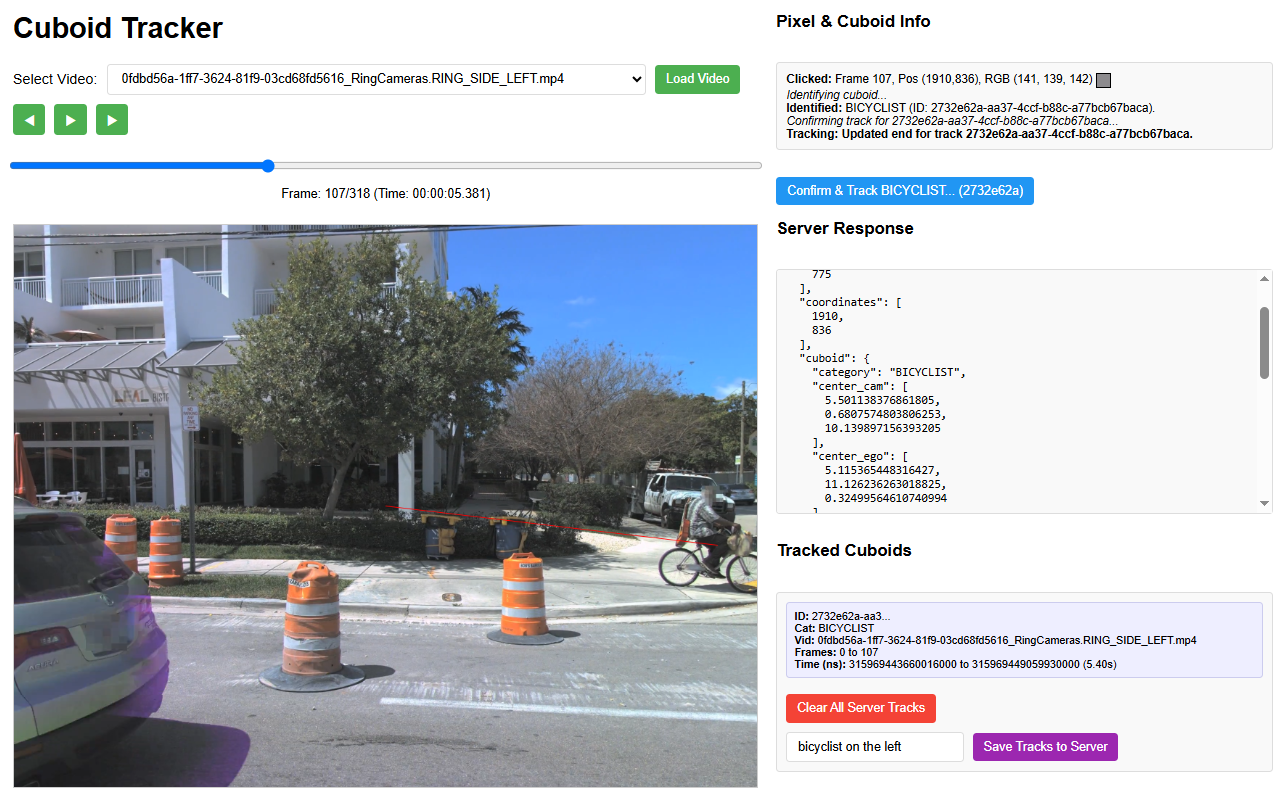}}
    }

    \caption{{\bf Manual Annotation Tool}. We create an annotation tool to assist with labeling manually defined scenarios. Our tool allows us to quickly annotate multi-object referential tracks in AV2.}
    \label{fig:annotation_tool}
\end{figure}

In order to manually annotate interesting scenarios, we build a custom web app with Claude 3.7 (Figure \ref{fig:annotation_tool}). To construct a scenario, the user selects an object by clicking on a point within the image frame, writes a natural language description, and selects the start and end frames of the \texttt{referred objects} correspond to the prompt. We project a ray originating from the 2D point and find the ground truth 3D bounding box centroid closest to the ray. This tool allows us to quickly generate multi-object and multi-camera referential tracks.

\section{Referential Tracking Evaluation of {\tt Related} and {\tt Other} Objects}
\label{sec:related_other}
RefProg creates a scene graph for each natural language prompt that not only identifies {\tt referred objects}, but also classifies tracks as {\tt related objects} and {\tt other objects}. For example, given the prompt {\tt vehicle near a pedestrian}, the vehicle would be the {\tt referred object}, the pedestrian would be the {\tt related object}, and everything else in the scene would be considered {\tt other objects}. We benchmark the base tracker with HOTA and evaluate HOTA-Temporal of {\tt related objects} and {\tt other objects} in Table \ref{tab:related_other}. Since most objects are not related to the prompt, HOTA and HOTA-Temporal {\tt Other} are similar. 

\begin{table}[h]
\centering
\caption{\textbf{Referential Tracking Accuracy for {\tt Related} and {\tt Other} Objects.} We present RefProg's referential tracking accuracy for {\tt related objects} and {\tt other objects}. We find that HOTA (for the base tracker) is similar to HOTA-Temporal ({\tt Other}) because most objects are not relevant to the referential prompt.}
\label{tab:related_other}
\resizebox{\linewidth}{!}{
\begin{tabular}{|l|c|c|c|}
\hline
\rowcolor{gray!10}
\textbf{Method} &
\textbf{HOTA $\uparrow$} &
\textbf{HOTA-Temporal ({\tt Related}) $\uparrow$} &
\textbf{HOTA-Temporal ({\tt Other}) $\uparrow$} \\
\hline

\rowcolor{gray!10}
\multicolumn{4}{|l|}{\bf Referential Tracking by Program Synthesis} \\
\hline

Ground Truth & 100.0 & 57.2 & 97.9 \\ \hline
Le3DE2E \cite{chenle3de2e} & 78.9 & 35.6 & 74.2 \\ \hline
ReVoxelDet \cite{lee2024revoxeldet} & 69.4 & 28.2 & 63.4 \\ \hline
TransFusion \cite{bai2022transfusion} & 69.5 & 28.6 & 63.8 \\ \hline
Valeo4Cast \cite{xu2024valeo4cast} & 75.1 & 30.8 & 69.2 \\ \hline

\end{tabular}
}
\end{table}

\section{Impact of Sampling Rate on RefProg}

We evaluate the impact of sampling rate on RefProg. First, we subsample three detectors (Le3DE2E \cite{chenle3de2e}, Valeo4Cast \cite{xu2024valeo4cast}, and BEVFusion \cite{liu2022bevfusion}) at 2Hz and 10 Hz. Next, we associate all detections with AB3DMOT \cite{weng20203d} to generate 2Hz and 10Hz tracks respectively. Lastly, we evaluate RefProg on all trackers at 2Hz. We find that standard tracking performance drops by 5\% when subsampling due to the difficulty of associating sparse detections. We see a similar drop in HOTA-Track, Log Balanced Accuracy and Timestamp Balanced Accuracy. Interestingly, we do not see a similar drop in performance with HOTA-Temporal. This suggests that RefProg might be robust to temporally sparse inputs and ID switches. 

\label{sec:sampling_rate}

\begin{table}[h]
\caption{\textbf{Impact of Sampling Rate.} We evaluate RefProg's performance with subsampled inputs to measure the relative impact on referential tracking accuracy. We find that standard tracking performance drops by 5\% when subsampling due to the difficulty of associating sparse detections.
} 
\label{tab:sampling}
\resizebox{\linewidth}{!}{
\begin{tabular}{|lccccccc|}
\hline
\rowcolor{gray!10}
\multicolumn{1}{|l|}{\textbf{Method}} & \multicolumn{1}{l|}{\textbf{Hz}} & \multicolumn{1}{l||}{\textbf{HOTA $\uparrow$}} & \multicolumn{1}{l|}{\textbf{HOTA-Temporal $\uparrow$}} & \multicolumn{1}{l|}{\textbf{HOTA-Track $\uparrow$}} & \multicolumn{1}{l|}{\textbf{Log Bal. Acc. $\uparrow$}} & \multicolumn{1}{l|}{\textbf{Timestamp Bal. Acc. $\uparrow$}}\\ \hline
\multicolumn{1}{|l|}{Ground Truth} & \multicolumn{1}{c|}{10}         & \multicolumn{1}{c||}{100.0}              & \multicolumn{1}{c|}{64.7}        & \multicolumn{1}{c|}{68.7}            &      \multicolumn{1}{c|}{81.1}     &      \multicolumn{1}{c|}{80.7}           \\ \hline
\multicolumn{1}{|l|}{Ground Truth} & \multicolumn{1}{c|}{2}         & \multicolumn{1}{c||}{100.0}              & \multicolumn{1}{c|}{64.2}        & \multicolumn{1}{c|}{68.8}            &      \multicolumn{1}{c|}{80.6}     &      \multicolumn{1}{c|}{80.1}           \\ \hline
\multicolumn{1}{|l|}{Le3DE2E Det. \cite{chenle3de2e} + AB3DMOT \cite{weng20203d}} & \multicolumn{1}{c|}{10}         & \multicolumn{1}{c||}{78.0}              & \multicolumn{1}{c|}{48.3}        & \multicolumn{1}{c|}{49.2}            &      \multicolumn{1}{c|}{73.1}     &      \multicolumn{1}{c|}{73.6}           \\ \hline
\multicolumn{1}{|l|}{Le3DE2E Det. \cite{chenle3de2e} + AB3DMOT \cite{weng20203d}} & \multicolumn{1}{c|}{2}         & \multicolumn{1}{c||}{73.0}              & \multicolumn{1}{c|}{45.2}        & \multicolumn{1}{c|}{44.1}            &      \multicolumn{1}{c|}{69.3}     &      \multicolumn{1}{c|}{69.5}           \\ \hline
\multicolumn{1}{|l|}{Valeo4Cast Det. \cite{xu2024valeo4cast} + AB3DMOT \cite{weng20203d}} & \multicolumn{1}{c|}{10}         & \multicolumn{1}{c||}{74.8}              & \multicolumn{1}{c|}{42.8}        & \multicolumn{1}{c|}{45.5}            &      \multicolumn{1}{c|}{71.6}     &      \multicolumn{1}{c|}{71.8}           \\ \hline
\multicolumn{1}{|l|}{Valeo4Cast Det. \cite{liu2022bevfusion} + AB3DMOT \cite{weng20203d}} & \multicolumn{1}{c|}{2}         & \multicolumn{1}{c||}{69.9}              & \multicolumn{1}{c|}{45.2}        & \multicolumn{1}{c|}{44.1}            &      \multicolumn{1}{c|}{69.3}     &      \multicolumn{1}{c|}{69.5}           \\ \hline
\multicolumn{1}{|l|}{BEVFusion Det. \cite{liu2022bevfusion} + AB3DMOT \cite{weng20203d}} & \multicolumn{1}{c|}{10}         & \multicolumn{1}{c||}{76.0}              & \multicolumn{1}{c|}{46.4}        & \multicolumn{1}{c|}{47.8}            &      \multicolumn{1}{c|}{73.7}     &      \multicolumn{1}{c|}{72.3}           \\ \hline
\multicolumn{1}{|l|}{BEVFusion Det. \cite{liu2022bevfusion} + AB3DMOT \cite{weng20203d}} & \multicolumn{1}{c|}{2}         & \multicolumn{1}{c||}{70.9}              & \multicolumn{1}{c|}{46.4}        & \multicolumn{1}{c|}{45.2}            &      \multicolumn{1}{c|}{72.1}     &      \multicolumn{1}{c|}{72.9}           \\ \hline
\end{tabular}}
\end{table}

\section{More Dataset Statistics}
\label{sec:dataset statistics}
\begin{figure}[t]
    \begin{minipage}{\linewidth}
        \begin{minipage}[c]{\linewidth}
            \centering
            \begin{tabular}[c]{|c|c|}
            \hline
            \rowcolor{gray!10}
            {\bf Expression Type} & {\bf \# Expressions} \\
            \hline
            Unique & 407 \\ \hline
            Positive Match & 7910 \\ \hline
            Negative Match & 2090 \\ \hline
            Procedurally Defined & 9880 \\ \hline
            Manually Defined & 120 \\  \hline
            Includes \texttt{Related Objects} & 5531  \\ \hline
            Ego is the only \texttt{Referred Object} & 827 \\ \hline
            Ego is the only \texttt{Related Object} & 1200 \\ \hline
            \end{tabular}
        \end{minipage} \\
        \begin{minipage}[c]{\linewidth}
            \centering
                \includegraphics[width=\linewidth]{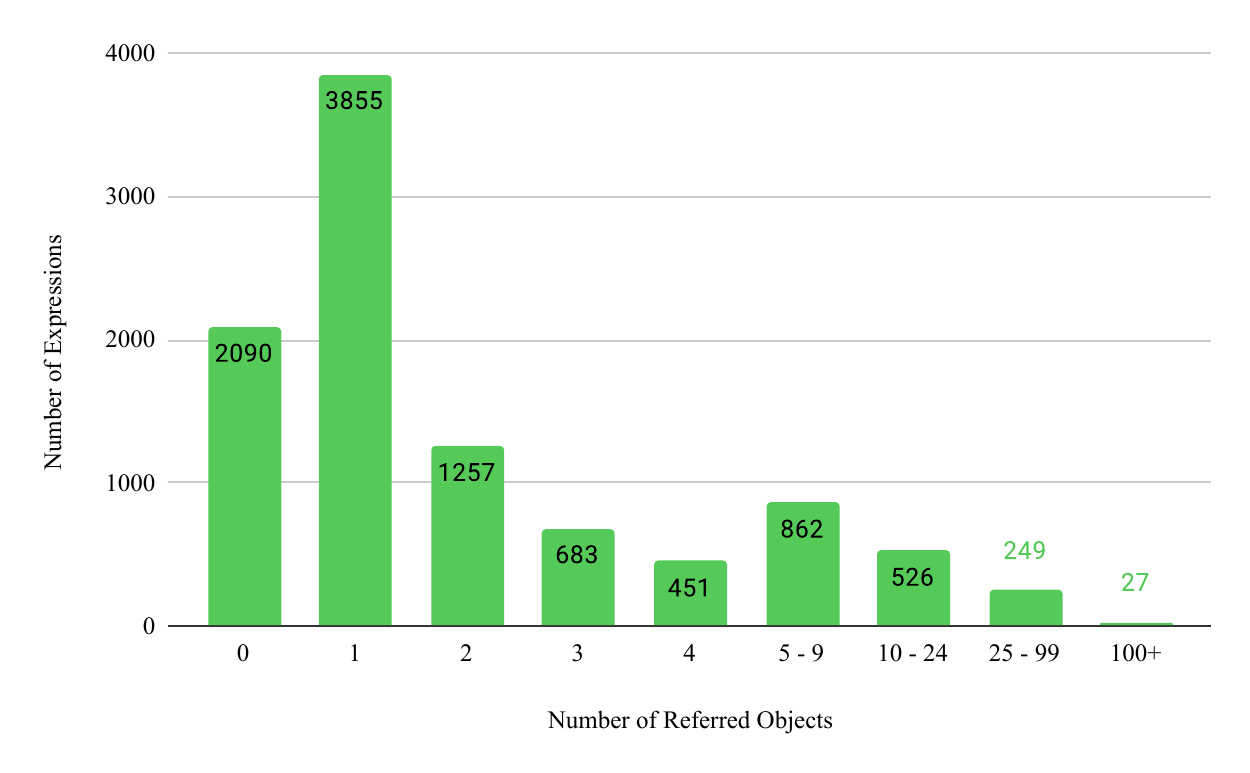}
        \end{minipage}
    \end{minipage}
    
    \caption{\textbf{RefAV Expression Statistics.} We compute the number of referring expressions categorized by expression type, count the number of the expressions that are positive or negative, and whether they are procedurally generated or manually constructed. We also highlight the number of positive expressions that include \texttt{related object} annotations and explicit references to the ego vehicle. The bar chart on the right shows the number of objects refered to by each expression. Most expressions in RefAV refer to zero (e.g. a negative match) or one object. On average, each expression refers to 3.92 objects.}
    \label{fig:refav_prompt_stats}
\end{figure}

\begin{figure}[t]
    \centering
    \includegraphics[width=\columnwidth,trim={0 1cm 0 0cm}]{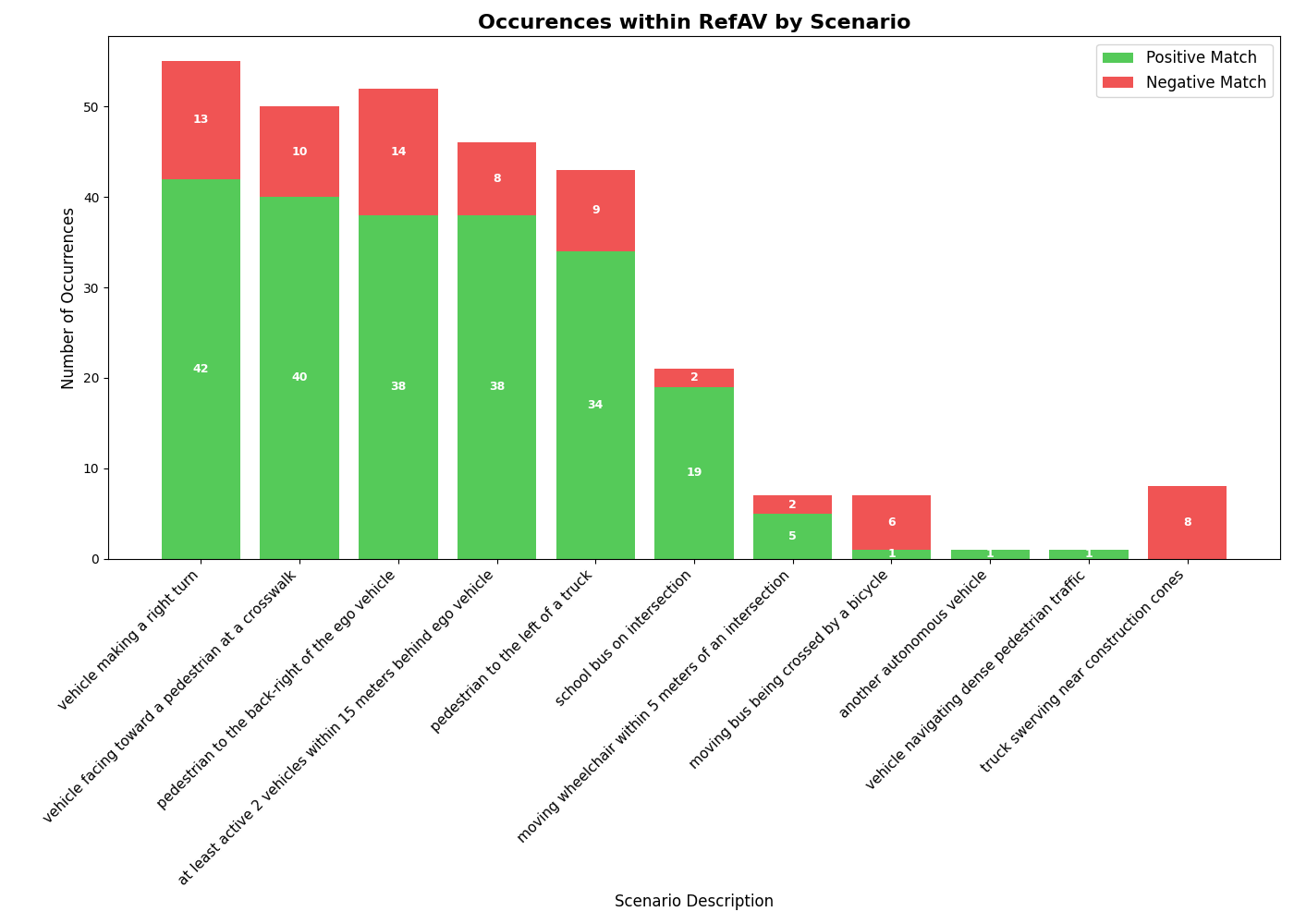}
    \caption{\textbf{Occurrences within RefAV by Scenario.} RefAV contains 407 unique scenario descriptions. We display 11 representative descriptions picked uniformly. After the diversity sampling process, RefAV includes about 40 positive matches for common scenarios such as {\tt vehicle making a right turn}. Less common scenarios such as {\tt moving bus being crossed by a bicycle} occur only in only a handful of logs across the ArgoVerse 2 dataset. RefAV includes negative examples for almost all procedurally defined scenarios. Manually annotated scenarios such as {\tt another autonomous vehicle} do not have corresponding negative examples.}
    \label{fig:refav_scenario_description_distribution}
\end{figure}

\begin{figure*}[t]
    \begin{minipage}{\linewidth}
        \begin{minipage}[c]{0.49\linewidth}
        \centering
        \includegraphics[width=\linewidth]{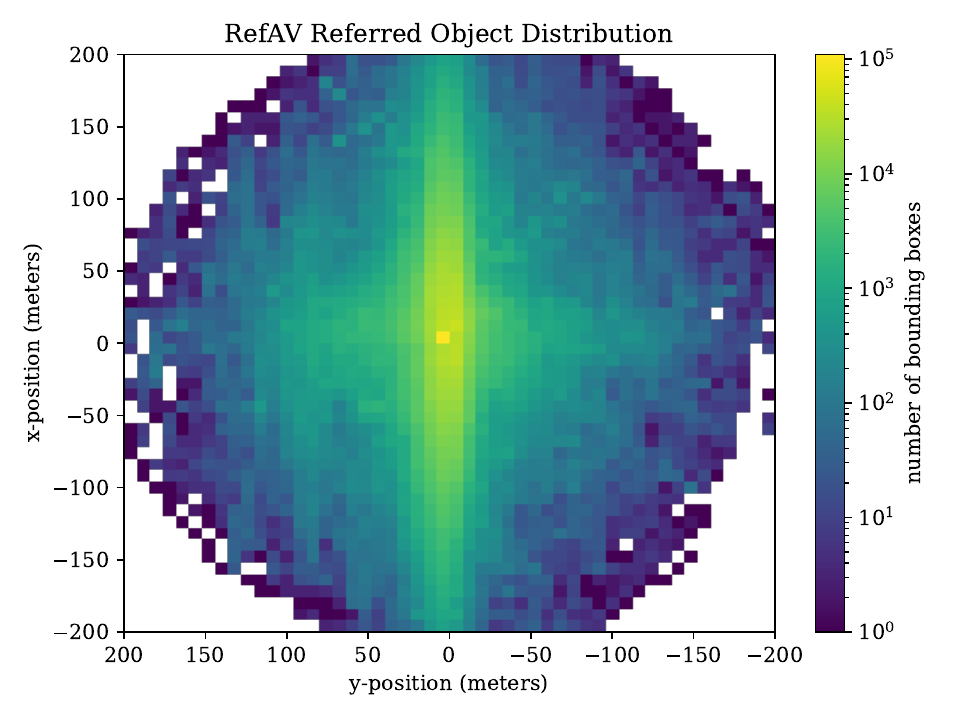}
        \end{minipage}
    \hfill
        \begin{minipage}[c]{0.49\linewidth}
            \centering
            
            \includegraphics[width=\linewidth]{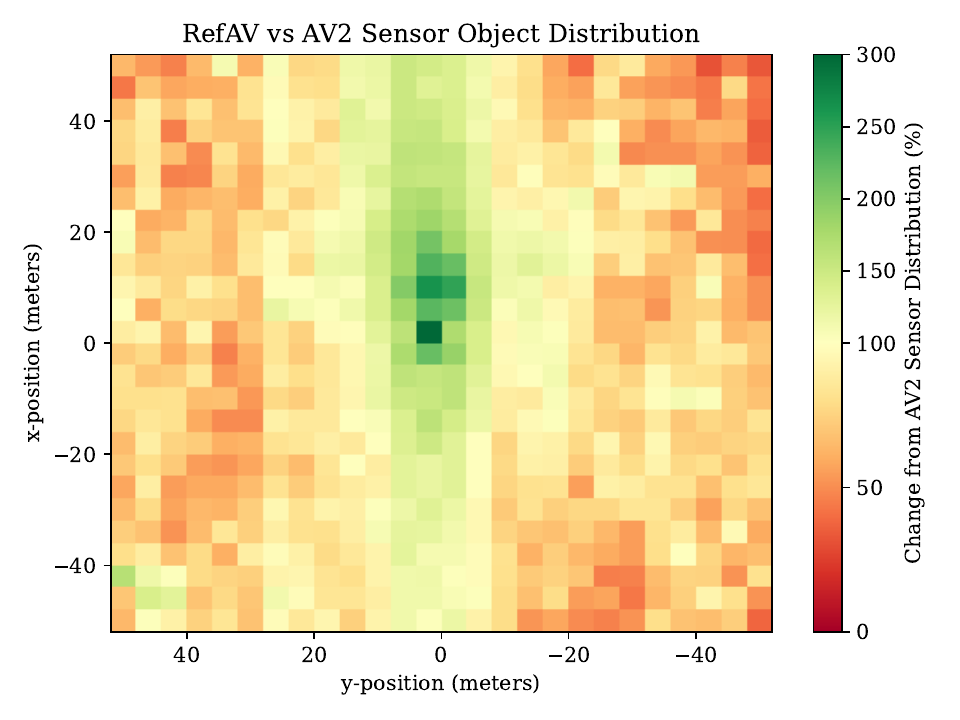}

        \end{minipage}
    \end{minipage}
    
    \caption{\textbf{RefAV Object Distribution.}  RefAV includes \texttt{referred objects} in all directions up to 150m away from the ego vehicle. RefAV places a special focus on the ego vehicle and objects that interact with the ego vehicle. Therefore, \texttt{referred objects} are disproportionately located on the road in front of the ego vehicle. Both {\tt referred object} heatmaps are in the ego vehicle coordinate frame.}
    \label{fig:refav_location_distribution}
\end{figure*}

 Figure \ref{fig:refav_prompt_stats} quantifies important characteristics of RefAV such as the number of unique scenario descriptions, the number of positive and negative examples, the number of procedurally and manually defined scenarios, the number of scenarios that include object relationships, and the number of scenarios directly involving the ego-vehicle. We use the catch-all term ``expression'' for scenario description to better compare RefAV to other datasets. A positive match occurs when the data log contains at least one {\tt referred object} corresponding to the scenario description. Figure \ref{fig:refav_scenario_description_distribution} shows the number of positive and negative examples for representative scenario descriptions. In Figure \ref{fig:refav_location_distribution}, we evaluate the spatial distribution of {\tt referred objects}.

\section{Summary of Competition Top Performers}
\label{sec:cvpr_results}

We summarize the contributions of top teams below. 

\textbf{Team Zeekr UMCV} extends RefProg with global context-aware generation to processes all queries collectively for improved reasoning and consistency. They also propose a multi-agent refinement system where a secondary refiner agent iteratively debugs and enhances the generated code.

\textbf{Team Mi3 UCM} proposes two key improvements to RefProg: (1) Fault-Tolerant Iterative Code Generation (FT-ICG), which refines faulty code by re-prompting the LLM with error feedback until successful execution, and (2) Enhanced Prompting for Spatial Relational Functions (EP-SRF), which clarifies function semantics to prevent misinterpretations of spatial relationships between objects.

\textbf{Team ZXH} refines RefProg by prompting LLMs with a batch of queries and in-context examples instead of prompting for each query one at a time. Importantly, this team's approach relies on in-context learning and batch prompting to improve consistency and efficiency.

\section{Examples of RefProg's Synthesized Programs}
We provide examples of RefProg’s correctly synthesized programs below.

\textit{Prompt: Pedestrian crossing between stopped buses}
\begin{lstlisting}[style=pythonstyle,language=Python]
# Description: pedestrian crossing between stopped buses

# First, let's get all pedestrians and buses from the dataset
pedestrians = get_objects_of_category(log_dir, category="PEDESTRIAN")
buses = get_objects_of_category(log_dir,  category="BUS")

# We need to find stopped buses
stationary_buses = stationary(buses, log_dir)

# Alternatively, we could use has_velocity with  a low threshold to find buses that are  temporarily stopped
stopped_buses = has_velocity(buses, log_dir, min_velocity=0, max_velocity=0.5)

# Now we need to find pedestrians that are crossing between the stopped buses. First, identify pedestrians that are crossing in front of a stopped bus
peds_crossing_bus = being_crossed_by(stationary_buses, pedestrians, log_dir)

# Reverse the relationship to get the pedestrians that are doing the crossing
crossing_peds = reverse_relationship(being_crossed_by)(stationary_buses, pedestrians, log_dir)

# Now we need to find pedestrians that have buses on both sides (left and right). This means the pedestrian is between buses
peds_with_bus_left = has_objects_in_relative_direction(crossing_peds, stationary_buses, log_dir, direction="left", min_number=1, within_distance=10)
peds_with_bus_right = has_objects_in_relative_direction(crossing_peds, stationary_buses, log_dir, direction="right", min_number=1, within_distance=10)

# Combine to find pedestrians with buses on both sides
peds_between_buses = scenario_and([peds_with_bus_left, peds_with_bus_right])

# Output the scenario
output_scenario(peds_between_buses, "pedestrian crossing between stopped buses", log_dir, output_dir)
\end{lstlisting}

\textit{Prompt: Group of at least 3 moving bicyclists with 5m from each other}
\begin{lstlisting}[style=pythonstyle,language=Python]
# Description: Group of at least 3 moving bicyclists within 5 meters of each other

# Get all bicyclists from the dataset
bicyclists = get_objects_of_category(log_dir, category="BICYCLIST")

# Filter out stationary bicyclists to get only moving ones
moving_bicyclists = scenario_not(stationary)(bicyclists, log_dir)

# Find bicyclists that have at least 2 other bicyclists within 5 meters (which makes a group of at least 3 bicyclists)
bicyclists_in_groups = near_objects(moving_bicyclists, moving_bicyclists, log_dir, distance_thresh=5, min_objects=2, include_self=False)

# Output the scenario
output_scenario(bicyclists_in_groups, "group of at least 3 moving bicyclists within 5 meters of each other", log_dir, output_dir)
\end{lstlisting}

\textit{Prompt: Accelerating vehicle changing lanes to the right}
\begin{lstlisting}[style=pythonstyle,language=Python]
# Description: accelerating vehicle changing lanes to the right
# First, get all vehicles in the scene
vehicles = get_objects_of_category(log_dir, category="VEHICLE")

# Find vehicles that are accelerating using the accelerating function with a reliable minimum acceleration threshold of 0.65
accelerating_vehicles = accelerating(vehicles, log_dir, min_accel=0.65)

# Find vehicles that are changing lanes to the right. We use the changing_lanes function with direction='right'
right_lane_changes = changing_lanes(vehicles, log_dir, direction='right')

# Now find the intersection of accelerating vehicles and those changing lanes to the right. This gives us vehicles that are both accelerating and changing lanes to the right
accelerating_and_changing_right = scenario_and([accelerating_vehicles, right_lane_changes])

# Output the scenario
output_scenario(accelerating_and_changing_right,
"accelerating vehicle changing lanes to the right",
log_dir, output_dir)
\end{lstlisting}

We provide examples of RefProg’s incorrectly synthesized programs below.

\textit{Prompt: bus in front of ego vehicle obstructing view of red traffic light}

The following program fails because the pre-defined atomic functions lack the expressivity to synthesize a correct program. Interestingly, the LLM realizes that the provided functions don’t allow us to determine if a traffic light is being obstructed. 

\begin{lstlisting}[style=pythonstyle,language=Python]
# Description: bus in front of ego vehicle obstructing view of red traffic light

# Get the ego vehicle and buses
ego_vehicle = get_objects_of_category(log_dir, category="EGO_VEHICLE")
buses = get_objects_of_category(log_dir, category="BUS")

# Find buses that are in front of the ego vehicle
# We need to find buses that are directly in front of the ego vehicle within a reasonable distance (20 meters) and with minimal lateral offset
buses_in_front_of_ego = get_objects_in_relative_direction(ego_vehicle, buses, log_dir, direction='forward', min_number=1, within_distance=20, lateral_thresh=3)

# We want buses that are moving in the same direction as the ego vehicle to ensure they're actually in the same lane and potentially blocking the view
buses_in_same_direction = heading_in_relative_direction_to(buses_in_front_of_ego, ego_vehicle, log_dir, direction='same')

# We need buses that are in the same lane as the ego vehicle
buses_in_same_lane = in_same_lane(buses_in_same_direction, ego_vehicle, log_dir)

# These buses are likely to be obstructing the view of traffic lights
# Note: The functions don't directly allow us to determine if a traffic light is being obstructed, but buses directly in front of the ego vehicle in the same lane are the most likely candidates for obstructing the view of traffic lights

output_scenario(buses_in_same_lane, "bus in front of ego vehicle obstructing view of red traffic light", log_dir, output_dir)
\end{lstlisting}

\textit{Prompt: bicycle following closely behind a vehicle}

The following program fails due to imprecise semantics because the bicycle is not necessarily following the vehicle.

\begin{lstlisting}[style=pythonstyle,language=Python]
# Description: bicycle following closely behind a vehicle
# We need to find bicycles that are following vehicles closely

# Get all bicycles and vehicles from the dataset
bicycles = get_objects_of_category(log_dir, category="BICYCLE")
vehicles = get_objects_of_category(log_dir, category="VEHICLE")

# We need to identify bicycles that are behind vehicles using get_objects_in_relative_direction to find bicycles that are behind vehicles. The bicycles should be within a reasonable distance (e.g., 10 meters) to be considered "following closely"
bicycles_behind_vehicles = get_objects_in_relative_direction(vehicles, bicycles, log_dir, direction="backward", within_distance=10, lateral_thresh=2)

# We might also want to ensure the bicycle is moving (not stationary)
moving_bicycles = scenario_not(stationary)(bicycles, log_dir)

# Combine the conditions: bicycles that are both moving and following closely behind vehicles
bicycles_following_vehicles = scenario_and([bicycles_behind_vehicles, moving_bicycles])

# Output the scenario
output_scenario(bicycles_following_vehicles, "bicycle following closely behind a vehicle", log_dir, output_dir)
\end{lstlisting}

\textit{Prompt: vehicle starting to switch lanes but then going back}

The following program uses incorrect logic. It fails to combine lane changes and acceleration and uses AND on exclusive sets. 

\begin{lstlisting}[style=pythonstyle,language=Python]
# Strategy:
# 1. Get all vehicles
# 2. Find vehicles that are changing lanes
# 3. Find vehicles that are also changing lanes in the opposite direction shortly after
# 4. This indicates a vehicle that started to change lanes but then went back

# Get all vehicles
vehicles = get_objects_of_category(log_dir, category="VEHICLE")

# Find vehicles that are moving (not stationary)
moving_vehicles = scenario_not(stationary)(vehicles, log_dir)

# Find vehicles that are changing lanes in either direction
lane_changing_vehicles = changing_lanes(moving_vehicles, log_dir)

# Find vehicles that are changing lanes to the left
left_lane_changes = changing_lanes(moving_vehicles, log_dir, direction='left')

# Find vehicles that are changing lanes to the right
right_lane_changes = changing_lanes(moving_vehicles, log_dir, direction='right')

# A vehicle that starts to switch lanes but goes back would show lateral acceleration in one direction followed by lateral acceleration in the opposite direction. We can look for vehicles with significant lateral acceleration in both directions

# Find vehicles with lateral acceleration to the left 
left_accel = has_lateral_acceleration(moving_vehicles, log_dir, min_accel=0.5)

# Find vehicles with lateral acceleration to the right 
right_accel = has_lateral_acceleration(moving_vehicles, log_dir, min_accel=-0.5, max_accel=0)

# Vehicles that have both left and right acceleration patterns are likely the ones that started to change lanes but went back
lane_change_aborted = scenario_and([left_accel, right_accel])
# Output the scenario
output_scenario(lane_change_aborted, description, log_dir, output_dir)
\end{lstlisting}

\section{API Listing}
\label{sec:api_listing}

We present our full Python API listing below. 
\begin{lstlisting}[style=pythonstyle,language=Python]
def has_objects_in_relative_direction(
    track_candidates:dict, 
    related_candidates:dict, 
    log_dir:Path, 
    direction:Literal["forward", "backward", "left", "right"], 
    min_number:int=1, 
    max_number:int=np.inf, 
    within_distance:float=50, 
    lateral_thresh:float=np.inf)-> dict:
    """
    Identifies tracked objects with at least the minimum number of related candidates in the specified direction. If the minimum number is met, will create relationships equal to the max_number of closest objects. 
    
    Args:
        track_candidates: Tracks to analyze (scenario dictionary). related_candidates: Candidates to check for in direction (scenario dictionary).
        log_dir: Path to scenario logs.
        direction: Direction to analyze from the track's point of view ('forward', 'backward', 'left', 'right').
        min_number: Minimum number of objects to identify in the direction per timestamp. Defaults to 1.
        max_number: Maximum number of objects to identify in the direction per timestamp. Defaults to infinity.
        within_distance: Maximum distance for considering an object in the direction. Defaults to infinity.
        lateral_thresh: Maximum lateral distance the related object can be from the sides of the tracked object. Defaults to infinity.
    
    Returns:
        dict: A scenario dictionary where keys are track UUIDs and values are dictionaries containing related candidate UUIDs and lists of timestamps when the condition is met for that relative direction.
    
    Example: 
        vehicles_with_peds_in_front = has_objects_in_relative_direction(vehicles, pedestrians, log_dir, direction='forward', min_number=2)
    """


def get_objects_in_relative_direction(
    track_candidates:dict, 
    related_candidates:dict, 
    log_dir:Path, 
    direction:Literal["forward", "backward", "left", "right"], 
    min_number:int=0, 
    max_number:int=np.inf, 
    within_distance:float=50, 
    lateral_thresh:float=np.inf)->dict:
    """
    Returns a scenario dictionary of the related candidates that are in the relative direction of the track candidates.
    
    
    Args:
        track_candidates: Tracks (scenario dictionary).
        related_candidates: Candidates to check for in direction (scenario dictionary).
        log_dir: Path to scenario logs.
        direction: Direction to analyze from the track's point of view ('forward', 'backward', 'left', 'right').
        min_number: Minimum number of objects to identify in the direction per timestamp. Defaults to 0.
        max_number: Maximum number of objects to identify in the direction per timestamp. Defaults to infinity.
        within_distance: Maximum distance for considering an object in the direction. Defaults to infinity.
        lateral_thresh: Maximum lateral distance the related object can be from the sides of the tracked object. Lateral distance is distance is the distance from the sides of the object that are parallel to the specified direction. Defaults to infinity.
    
    Returns:
        dict: A scenario dictionary where keys are track UUIDs and values are dictionaries containing related candidate UUIDs and lists of timestamps when the condition is met for that relative direction.
    
    Example: 
        peds_in_front_of_vehicles = get_objects_in_relative_direction(vehicles, pedestrians, log_dir, direction='forward', min_number=2)
    """


def get_objects_of_category(
    log_dir, 
    category)->dict:
    """
    Returns all objects from a given category from the log annotations. This method accepts the super-categories "ANY" and "VEHICLE".
    
    Args:
        log_dir: Path to the directory containing scenario logs and data. 
        category: the category of objects to return
    
    Returns: 
        dict: A scenario dict that where keys are the unique id (uuid) of the object and values are the list of timestamps the object is in view of the ego-vehicle.
    
    Example: 
        trucks = get_objects_of_category(log_dir, category='TRUCK')
    """


def is_category(
    track_candidates:dict, 
    log_dir:Path, 
    category:str)->dict:
    """
    Returns all objects from a given category from track_candidates dict. This method accepts the super-categories "ANY" and "VEHICLE".
    
    Args:
        track_candidates: The scenario dict containing the objects to filter down
        log_dir: Path to the directory containing scenario logs and data. 
        category: the category of objects to return
    
    Returns: 
        dict: A scenario dict that where keys are the unique id of the object of the given category and values are the list of timestamps the object is in view of the ego-vehicle.
    
    Example: 
        box_trucks = is_category(vehicles, log_dir, category='BOX_TRUCK')
    """


def turning(
    track_candidates: dict, 
    log_dir:Path, 
    direction:Literal["left", "right", None]=None)->dict:
    """
    Returns objects that are turning in the given direction. 
    
    Args:
        track_candidates: The objects you want to filter from (scenario dictionary).
        log_dir: Path to scenario logs.
        direction: The direction of the turn, from the track's point of view ('left', 'right', None).
    
    Returns:
        dict: A filtered scenario dictionary where keys are track UUIDs that meet the turning criteria and values are nested dictionaries containing timestamps.
    
    Example: 
        turning_left = turning(vehicles, log_dir, direction='left')
    """


def changing_lanes(
    track_candidates:dict, 
    log_dir:Path, 
    direction:Literal["left", "right", None]=None)-> dict:
    """
    Identifies lane change events for tracked objects in a scenario.
    
    Args:
        track_candidates: The tracks to analyze (scenario dictionary).
        log_dir: Path to scenario logs.
        direction: The direction of the lane change. None indicates tracking either left or right lane changes ('left', 'right', None).
    
    Returns:
        dict: A filtered scenario dictionary where keys are track UUIDs that meet the lane change criteria and values are nested dictionaries containing timestamps and related data.
    
    Example: 
        left_lane_changes = changing_lanes(vehicles, log_dir, direction='left')
    """


def has_lateral_acceleration(
    track_candidates:dict, 
    log_dir:Path, 
    min_accel=-np.inf, 
    max_accel=np.inf)-> dict:
    """
    Objects with a lateral acceleartion between the minimum and maximum thresholds. Most objects with a high lateral acceleration are turning. Postive values indicate accelaration to the left while negative values indicate acceleration to the right. 
    
    Args:
        track_candidates: The tracks to analyze (scenario dictionary).
        log_dir: Path to scenario logs.
        direction: The direction of the lane change. None indicates tracking either left or right lane changes ('left', 'right', None).
    
    Returns:
        dict: A filtered scenario dictionary where keys are track UUIDs that meet the lane change criteria and values are nested dictionaries containing timestamps and related data.
    
    Example: 
        jerking_left = has_lateral_acceleration(non_turning_vehicles, log_dir, min_accel=2)
    """


def facing_toward(
    track_candidates:dict,
    related_candidates:dict,
    log_dir:Path,
    within_angle:float=22.5,
    max_distance:float=50)->dict:
    """
    Identifies objects in track_candidates that are facing toward objects in related candidates. The related candidate must lie within a region lying within within_angle degrees on either side the track-candidate's forward axis.
    
    Args:
        track_candidates: The tracks that could be heading toward another tracks
        related_candidates: The objects to analyze to see if the track_candidates are heading toward
        log_dir:  Path to the directory containing scenario logs and data.
        fov: The field of view of the track_candidates. The related candidate must lie within a region lying within fov/2 degrees on either side the track-candidate's forward axis.
        max_distance: The maximum distance a related_candidate can be away to be considered by 
    
    Returns:
        dict: A filtered scenario dict that contains the subset of track candidates heading toward at least one of the related candidates.
    
    Example: 
        pedestrian_facing_away = scenario_not(facing_toward)(pedestrian, ego_vehicle, log_dir, within_angle=180)
    """


def heading_toward(
    track_candidates:dict,
    related_candidates:dict,
    log_dir:Path,
    angle_threshold:float=22.5,
    minimum_speed:float=.5,
    max_distance:float=np.inf)->dict:
    """
    Identifies objects in track_candidates that are heading toward objects in related candidates. The track candidates acceleartion vector must be within the given angle threshold of the relative position vector. The track candidates must have a component of velocity toward the related candidate greater than the minimum_accel.
    
    Args:
        track_candidates: The tracks that could be heading toward another tracks
        related_candidates: The objects to analyze to see if the track_candidates are heading toward
        log_dir:  Path to the directory containing scenario logs and data.
        angle_threshold: The maximum angular difference between the velocity vector and relative position vector between the track candidate and related candidate.
        min_vel: The minimum magnitude of the component of velocity toward the related candidate
        max_distance: Distance in meters the related candidates can be away from the track candidate to be considered
    
    Returns:
        dict: A filted scenario dict that contains the subset of track candidates heading toward at least one of the related candidates.
    
    Example: 
        heading_toward_traffic_cone = heading_toward(vehicles, traffic_cone, log_dir)
    """


def accelerating(
    track_candidates:dict,
    log_dir:Path,
    min_accel:float=.65,
    max_accel:float=np.inf)->dict:
    """
    Identifies objects in track_candidates that have a forward acceleration above a threshold. Values under -1 reliably indicates braking. Values over 1.0 reliably indiciates accelerating.
    
    Args:
        track_candidates: The tracks to analyze for acceleration (scenario dictionary)
        log_dir:  Path to the directory containing scenario logs and data.
        min_accel: The lower bound of acceleration considered
        max_accel: The upper bound of acceleration considered
    
    Returns:
        dict: A filtered scenario dictionary containing the objects with an acceleration between the lower and upper bounds.
    
    Example: 
        accelerating_motorcycles = accelerating(motorcycles, log_dir)
    """


def has_velocity(
    track_candidates:dict,
    log_dir:Path,
    min_velocity:float=.5,
    max_velocity:float=np.inf)->dict:
    """
    Identifies objects with a velocity between the given maximum and minimum velocities in m/s. Stationary objects may have a velocity up to 0.5 m/s due to annotation jitter.
    
    Args:
        track_candidates: Tracks to analyze (scenario dictionary).
        log_dir: Path to scenario logs.
        min_velocity: Minimum velocity (m/s). Defaults to 0.5.
        max_velocity: Maximum velocity (m/s)
    
    Returns:
        dict: Filtered scenario dictionary of objects meeting the velocity criteria.
    
    Example: 
        fast_vehicles = has_min_velocity(vehicles, log_dir, min_velocity=5)
    """


def at_pedestrian_crossing(
    track_candidates:dict,
    log_dir:Path,
    within_distance:float=1)->dict:
    """
    Identifies objects that within a certain distance from a pedestrian crossing. A distance of zero indicates that the object is within the boundaries of the pedestrian crossing.
    
    Args:
        track_candidates: Tracks to analyze (scenario dictionary).
        log_dir: Path to scenario logs.
        within_distance: Distance in meters the track candidate must be from the pedestrian crossing. A distance of zero means that the object must be within the boundaries of the pedestrian crossing.
    
    Returns:
        dict: Filtered scenario dictionary where keys are track UUIDs and values are lists of timestamps.
    
    Example: 
        vehicles_at_ped_crossing = at_pedestrian_crossing(vehicles, log_dir)
    """


def on_lane_type(
    track_uuid:dict,
    log_dir,
    lane_type:Literal["BUS", "VEHICLE", "BIKE"])->dict:
    """
    Identifies objects on a specific lane type.
    
    Args:
        track_candidates: Tracks to analyze (scenario dictionary).
        log_dir: Path to scenario logs.
        lane_type: Type of lane to check ('BUS', 'VEHICLE', or 'BIKE').
    
    Returns:
        dict: Filtered scenario dictionary where keys are track UUIDs and values are lists of timestamps.
    
    Example: 
        vehicles_on_bus_lane = on_lane_type(vehicles, log_dir, lane_type="BUS")
    """


def near_intersection(
    track_uuid:dict,
    log_dir:Path,
    threshold:float=5)->dict:
    """
    Identifies objects within a specified threshold of an intersection in meters.
    
    Args:
        track_candidates: Tracks to analyze (scenario dictionary).
        log_dir: Path to scenario logs.
        threshold: Distance threshold (in meters) to define "near" an intersection.
    
    Returns:
        Filtered scenario dictionary where keys are track UUIDs and values are lists of timestamps.
    
    Example: 
        bicycles_near_intersection = near_intersection(bicycles, log_dir, threshold=10.0)
    """


def on_intersection(
    track_candidates:dict, 
    log_dir:Path)->dict:
    """
    Identifies objects located on top of an road intersection.
    
    Args:
        track_candidates: Tracks to analyze (scenario dictionary).
        log_dir: Path to scenario logs.
    
    Returns:
        dict: Filtered scenario dictionary where keys are track UUIDs and values are lists of timestamps.
    
    Example: 
        strollers_on_intersection = on_intersection(strollers, log_dir)
    """


def being_crossed_by(
    track_candidates:dict,
    related_candidates:dict,
    log_dir:Path,
    direction:Literal["forward", "backward", "left", 
    "right"]="forward",
    in_direction:Literal['clockwise','counterclockwise',
    'either']='either',
    forward_thresh:float=10,
    lateral_thresh:float=5)->dict:
    """
    Identifies objects that are being crossed by one of the related candidate objects. A crossing is defined as the related candidate's centroid crossing the half-midplane of a tracked candidate. The direction of the half-midplane is specified with the direction. 
    
    Args:
        track_candidates: Tracks to analyze .
        related_candidates: Candidates (e.g., pedestrians or vehicles) to check for crossings.
        log_dir: Path to scenario logs.
        direction: specifies the axis and direction the half midplane extends from 
        in_direction: which direction the related candidate has to cross the midplane for it to be considered a crossing
        forward_thresh: how far the midplane extends from the edge of the tracked object
        lateral_thresh: the two planes offset from the midplane. If an related candidate crosses the midplane, it will continue being considered crossing until it goes past the lateral_thresh.
    
    Returns:
        dict: A filtered scenario dictionary containing all of the track candidates that were crossed by the related candidates given the specified constraints.
    
    Example: 
        overtaking_on_left = being_crossed_by(moving_cars, moving_cars, log_dir, direction="left", in_direction="clockwise", forward_thresh=4)
        vehicles_crossed_by_peds = being_crossed_by(vehicles, pedestrians, log_dir)
    """


def near_objects(
    track_uuid:dict,
    candidate_uuids:dict,
    log_dir:Path,
    distance_thresh:float=10,
    min_objects:int=1,
    include_self:bool=False)->dict:
    """
    Identifies timestamps when a tracked object is near a specified set of related objects.
    
    Args:
        track_candidates: Tracks to analyze (scenario dictionary).
        related_candidates: Candidates to check for proximity (scenario dictionary).
        log_dir: Path to scenario logs.
        distance_thresh: Maximum distance in meters a related candidate can be away to be considered "near".
        min_objects: Minimum number of related objects required to be near the tracked object.
    
    Returns:
        dict: A scenario dictionary where keys are timestamps when the tracked object is near the required number of related objects and values are lists of related candidate UUIDs present at those timestamps.
    
    Example: 
        vehicles_near_ped_group = near_objects(vehicles, pedestrians, log_dir, min_objects=3)
    """


def following(
    track_uuid:dict,
    candidate_uuids:dict,
    log_dir:Path)-> dict:
    """
    Returns timestamps when the tracked object is following a lead object. Following is defined simultaneously moving in the same direction and lane.
    """


def heading_in_relative_direction_to(
    track_candidates, 
    related_candidates, 
    log_dir, 
    direction:Literal['same', 'opposite', 'perpendicular'])->dict:
    """
    Returns the subset of track candidates that are traveling in the given direction compared to the related candidates.
    
    Arguments:
        track_candidates: The set of objects that could be traveling in the given direction
        related_candidates: The set of objects that the direction is relative to
        log_dir: The path to the log data
        direction: The direction that the positive tracks are 
        traveling in relative to the related candidates:
            - "opposite" indicates the track candidates are traveling in a direction 135-180 degrees from the direction the related candidates are heading toward.
            - "same" indicates the track candidates that are traveling in a direction 0-45 degrees from the direction the related candiates are heading toward.
            - "perpendicular" indicates the track candidates that are traveling in a direction 45-135 degrees from the direction the related candiates are heading toward.
    
    Returns:
        dict: the subset of track candidates that are traveling in the given direction compared to the related candidates.
    
    Example: 
        oncoming_traffic = heading_in_relative_direction_to(vehicles, ego_vehicle, log_dir, direction='opposite')    
    """


def stationary(
    track_candidates:dict, 
    log_dir:Path)->dict:
    """
    Returns objects that moved less than 2m over their length of observation in the scneario. This object is only intended to separate parked from active vehicles. Use has_velocity() with thresholding if you want to indicate vehicles that are temporarily stopped.
    
    Args:
        track_candidates: Tracks to analyze (scenario dictionary).
        log_dir: Path to scenario logs.
    
    Returns:
        dict: A filtered scenario dictionary where keys are track UUIDs and values are lists of timestamps when the object is stationary.
    
    Example: 
        parked_vehicles = stationary(vehicles, log_dir)
    """


def at_stop_sign(
    track_candidates:dict, 
    log_dir:Path, 
    forward_thresh:float=10)->dict:
    """
    Identifies timestamps when a tracked object is in a lanee corresponding to a stop sign. The tracked object must be within 15m of the stop sign. This may highlight vehicles using street parking near a stopped sign.
    
    Args:
        track_candidates: Tracks to analyze (scenario dictionary).
        log_dir: Path to scenario logs.
        forward_thresh: Distance in meters the vehcile is from the stop sign in the stop sign's front direction
    
    Returns:
        dict: A filtered scenario dictionary where keys are track UUIDs and values are lists of timestamps when the object is at a stop sign.
    
    Example: 
        vehicles_at_stop_sign = at_stop_sign(vehicles, log_dir)
    """


def in_drivable_area(
    track_candidates:dict, 
    log_dir:Path)->dict:
    """
    Identifies objects within track_candidates that are within a drivable area.
    
    Args:
        track_candidates: Tracks to analyze (scenario dictionary).
        log_dir: Path to scenario logs.
    
    Returns:
        dict: A filtered scenario dictionary where keys are track UUIDs and values are lists of timestamps when the object is in a drivable area.
    
    Example: 
        buses_in_drivable_area = in_drivable_area(buses, log_dir)
    """


def on_road(
    track_candidates:dict,
    log_dir:Path)->dict:
    """
    Identifies objects that are on a road or bike lane. This function should be used in place of in_driveable_area() when referencing objects that are on a road. The road does not include  parking lots or other driveable areas connecting the road to parking lots.
    
    Args:
        track_candidates: Tracks to filter (scenario dictionary).
        log_dir: Path to scenario logs.
    
    Returns:
        dict: The subset of the track candidates that are currently on a road.
    
    Example: 
        animals_on_road = on_road(animals, log_dir)   
    """


def in_same_lane(
    track_candidates:dict,
    related_candidates:dict,
    log_dir:Path)->dict:
    """
    Identifies tracks that are in the same road lane as a related candidate. 
    
    Args:
        track_candidates: Tracks to filter (scenario dictionary)
        related_candidates: Potential objects that could be in the same lane as the track (scenario dictionary)
        log_dir: Path to scenario logs.
    
    Returns:
        dict: A filtered scenario dictionary where keys are track UUIDs and values are lists of timestamps when the object is on a road lane.
    
    Example: 
        bicycle_in_same_lane_as_vehicle = in_same_lane(bicycle, regular_vehicle, log_dir)    
    """

def on_relative_side_of_road(
    track_candidates:dict, 
    related_candidates:dict, 
    log_dir:Path, 
    side=Literal['same', 'opposite'])->dict:
    """
    Identifies tracks that are in the same road lane as a related candidate. 
    
    Args:
        track_candidates: Tracks to filter (scenario dictionary)
        related_candidates: Potential objects that could be in the same lane as the track (scenario dictionary)
        log_dir: Path to scenario logs.
    
    Returns:
        dict: A filtered scenario dictionary where keys are track UUIDs and values are lists of timestamps when the object is on a road lane.
    
    Example: 
        bicycle_in_same_lane_as_vehicle = in_same_lane(bicycle, regular_vehicle, log_dir)    
    """

def is_color(
    track_candidates: dict, 
    log_dir: Path, 
    color:Literal["white", "silver", "black", "red", "yellow", "blue"])->dict:
    """
    Returns objects that are the given color, determined by SIGLIP2 feature similarity.

    Args:
        track_candidates: The objects you want to filter from (scenario dictionary).
        log_dir: Path to scenario logs.
        color: The color of the objects you want to return. Must be one of 'white', 'silver', 'black', 'red', 'yellow', or 'blue'. Inputting a different color defaults to returning all objects.

    Returns:
        dict: A filtered scenario dictionary where keys are track UUIDs that meet the turning criteria and values are nested dictionaries containing timestamps.

    Example: 
        red_cars = is_color(cars, log_dir, color='red')
    """

def scenario_and(scenario_dicts)->dict:
    """
    Returns a composed scenario where the track objects are the intersection of all of the track objects with the same uuid and timestamps.
    
    Args:
        scenario_dicts: the scenarios to combine 
    
    Returns:
        dict: a filtered scenario dictionary that contains tracked objects found in all given scenario dictionaries
    
    Example: 
        jaywalking_peds = scenario_and([peds_on_road, peds_not_on_pedestrian_crossing])
    """


def scenario_or(scenario_dicts)->dict:
    """
    Returns a composed scenario where that tracks all objects and relationships in all of the input scenario dicts.
    
    Args:
        scenario_dicts: the scenarios to combine 
    
    Returns:
        dict: an expanded scenario dictionary that contains every tracked object in the given scenario dictionaries
    
    Example: 
        be_cautious_around = scenario_or([animal_on_road, stroller_on_road])
    """


def reverse_relationship(func)->dict:
    """
    Wraps relational functions to switch the top level tracked objects and relationships formed by the function. 
    
    Args:
        relational_func: Any function that takes track_candidates and related_candidates as its first and second arguements
    
    Returns:
        dict: scenario dict with swapped top-level tracks and related candidates
    
    Example: 
        group_of_peds_near_vehicle = reverse_relationship(near_objects)(vehicles, peds, log_dir, min_objects=3)
    """


def scenario_not(func)->dict:
    """
    Wraps composable functions to return the difference of the input track dict and output scenario dict. Using scenario_not with a composable relational function will not return any relationships. 
    
    Args:
        composable_func: Any function that takes track_candidates as its first input
    
    Returns:
        dict: scenario dict with tracks that are not filtered by func
        
    Example: 
        active_vehicles = scenario_not(stationary)(vehicles, log_dir)
    """


def output_scenario(
    scenario:dict, 
    description:str, 
    log_dir:Path, 
    output_dir:Path, 
    visualize:bool=False, 
    **visualization_kwargs):
    """
    Outputs a file containing the predictions in an evaluation-ready format. Do not provide any visualization kwargs. 
    """

\end{lstlisting}

\end{document}